\documentclass[journal]{IEEEtran}
\IEEEoverridecommandlockouts
\usepackage{cite}
\usepackage{amsmath,amssymb,amsfonts}
\usepackage{graphicx}
\usepackage{textcomp}
\usepackage{xcolor}
\usepackage{amsmath,amsfonts}
\usepackage{amsthm}
\theoremstyle{remark}
\newtheorem{theorem}{Theorem}

\newtheorem{remark}{Remark}

\usepackage{array}
\usepackage{textcomp}
\usepackage{stfloats}
\usepackage{adjustbox}
\usepackage{multirow}
\usepackage{diagbox} 
\usepackage{tikz} 
\usepackage{url}
\usepackage{verbatim}
\usepackage{graphicx}
\usepackage{cite}
\usepackage{tabularx}
\usepackage[caption=false,font=footnotesize,labelfont=rm,textfont=rm]{subfig}
\usepackage{float}     
\usepackage{multirow}
\usepackage{booktabs}
\usepackage{colortbl}
\usepackage{xcolor}
\usepackage[linesnumbered, ruled]{algorithm2e}
\usepackage{hyperref}
\def\BibTeX{{\rm B\kern-.05em{\sc i\kern-.025em b}\kern-.08em
    T\kern-.1667em\lower.7ex\hbox{E}\kern-.125emX}}
\begin{document}

\title{Federated 3D Gaussian Splatting for Large-Scale Scene Reconstruction at Wireless Edge \\

}

\author{Guanlin Wu, Chao Hu, Pu Chen, Juyong Zhang, Han Hu, Shuguang Cui, \textit{Fellow, IEEE}, and Jie Xu, \textit{Fellow, IEEE}

\vspace{-0.4cm}

\thanks{{Part of this paper was presented at the IEEE Wireless Commun. Netw. Conf. Workshops (WCNCW), Kuala Lumpur, Malaysia, in Apr. 2026 \cite{Wu2026WCNC}.} }
\thanks{G. Wu, C. Hu, S. Cui, and J. Xu are with the School of Science and Engineering, the Shenzhen Future Network of Intelligence Institute, and the Guangdong Provincial Key Laboratory of Future Networks of Intelligence, The Chinese University of Hong Kong (Shenzhen), Longgang, Shenzhen 518172, China (e-mail: guanlinwu1@link.cuhk.edu.cn, chaohu@link.cuhk.edu.cn, shuguangcui@cuhk.edu.cn, xujie@cuhk.edu.cn). J. Xu is the corresponding author.}%
\thanks{P. Chen is with the Department of Computer Science, University of Liverpool, Liverpool, L697ZX, UK (e-mail: sgpchen4@liverpool.ac.uk).}

\thanks{J. Zhang is with the School of Mathematical Science, University of Science and Technology of China, Hefei 230026, China (e-mail: juyong@ustc.edu.cn).}
\thanks{H. Hu is with the School of Information and Electronics, Beijing Institute
of Technology, Beijing 100081, China (e-mail: hhu@bit.edu.cn).}
}

\maketitle

\begin{abstract}
  Three-dimensional (3D) Gaussian splatting (3D-GS) has emerged as a promising technique for large-scale scene reconstruction due to its high rendering efficiency and fidelity. However, the training of large-scale 3D-GS models at wireless edge faces various technical challenges including the limited communication, computation, and graphics processing unit (GPU) memory resources at edge devices, the structural inconsistency issue across local models hindering their effective aggregation, as well as privacy leakage risks associated with raw visual content and camera parameters. To address these challenges, this paper proposes a novel resource-efficient federated learning framework for efficiently training 3D-GS models of large scenes under severe resource constraints. First, we propose an on-device model lightweighting mechanism that adaptively selects and prunes Gaussian points to balance the rendering quality and training efficiency. In this mechanism, we quantitatively evaluate the importance of different Gaussian points at each device to facilitate the pruning, and use a novel importance-to-latency ratio criterion to determine the number of pruned Gaussian points under GPU memory and computation/communication latency constraints. Furthermore, we develop a 3D-GS model recovery mechanism that restores structural consistency across local 3D-GS models without accessing private camera parameters, enabling their effective aggregation towards a global model. Finally, extensive experiments show that our approach significantly accelerates convergence, maintains high rendering quality, and reduces training latency compared to state-of-the-art federated 3D-GS baselines.
\end{abstract}


\section{Introduction}

Large-scale three-dimensional (3D) scene reconstruction has recently garnered rapidly growing interest, which aims at generating high-fidelity 3D models of large-scale real-world environments from multi-view two-dimensional (2D) images captured from disparate viewpoints. By enabling accurate and immersive representations of complex settings, this technology plays a critical role in facilitating a wide range of advanced applications, such as extended reality \cite{9757457}, autonomous driving \cite{Zhou_2024_CVPR}, and embodied intelligence \cite{10930696}. Among various 3D reconstruction techniques, 3D Gaussian splatting (3D-GS) has emerged as a highly promising approach for large-scale scene reconstruction \cite{3DGS}. Its strength lies in representing scenes using a collection of learnable 3D Gaussian primitives, which facilitate highly efficient, real-time rendering while achieving photorealistic quality, significantly surpassing conventional methods in both rendering speed and visual fidelity \cite{survey1,chen2025survey3dgaussiansplatting}. {Recent works further improve the reconstruction quality of 3D-GS, for instance, by leveraging a neural radiance field (NeRF) as an assistant to guide Gaussian optimization \cite{nerfgs}.} However, as the scene scale increases, 3D-GS models grow substantially in size, introducing considerable challenges in graphics processing unit (GPU) memory consumption, training latency, and computational overhead. {To alleviate these issues, several works compress 3D-GS by evaluating the importance of individual Gaussian points and pruning the redundant points \cite{lightgs, minisplat}.}

Driven by advancements in the Artificial Intelligence of Things (AIoT) and wireless networks, a multitude of spatially distributed devices such as autonomous vehicles, virtual reality (VR) headsets, smartphones, and smart cameras are now equipped with enhanced sensing, communication, computation, and artificial intelligence (AI) capabilities. These devices can collectively capture multi-view 2D images for 3D-GS model reconstruction. Conventionally, the training of large-scale 3D-GS models relies on cloud computing, requiring distributed devices to transmit raw image data to a centralized data center. This approach, however, introduces significant challenges, including communication bottlenecks, privacy leakage risks, and under-utilization of the growing computational power available at the network edge \cite{Zhu2023, 9606720}. Consequently, instead of centralized training, distributed large-scale 3D-GS training has emerged as a promising alternative paradigm. In this paradigm, distributed nodes leverage their collective communication and computational resources to collaborate on training 3D-GS models while preserving data privacy \cite{WUembracing24}.

In the literature, only a handful of prior works have explored distributed 3D-GS training for large-scale scene reconstruction \cite{VastGaussian, DOGS,cosurfgs,dgtr,radiant,fed3Dgs}. First, some works focused on dividing a large-scale scene into smaller sub-scenes, enabling each distributed device to handle the computational load associated with training a sub-scene independently \cite{VastGaussian, DOGS}. In \cite{VastGaussian}, the authors proposed VastGaussian, a distributed 3D-GS training framework that utilizes the position and the visibility of the camera to divide the large-scale scene for training. In \cite{DOGS}, the authors adopted a recursive scene partitioning strategy to balance the computational workload across distributed devices. {Similarly, CityGaussian \cite{citygs} adopts a divide-and-conquer training scheme together with a level-of-detail strategy to enable real-time rendering of city-scale scenes.} Next, some other works investigated how to merge the local models trained by different devices into a complete large global model \cite{cosurfgs, dgtr}. These approaches merge all local models on a central node and perform global model refinement via model distillation to mitigate boundary artifact points. Furthermore, to enhance system scalability and flexibility, the authors in \cite{radiant} further investigated hierarchical cloud–edge–device training architectures, enabling centralized coordination through a small number of intermediate edge nodes. In another line of research, federated learning (FL) has recently emerged as a privacy-preserving and communication-efficient solution for 3D-GS training at the wireless edge \cite{fed3Dgs}. By allowing edge devices to perform on-device training and exchange only model updates, FL enables efficient coordination and accurate scene reconstruction without transmitting raw multi-view data to the server. {In summary, compared to centralized 3D-GS training that gathers all raw multi-view data at a single node (e.g., cloud and edge server), the above distributed and federated learning approaches instead train models over communication rounds. The two paradigms trade off differently: centralized raw-data upload incurs a single large transmission and concentrates the full training workload on the server, whereas repeated model synchronization spreads communication across rounds while exploiting distributed on-device computation. The latter is more favorable in the large-scale edge scenarios considered in this paper, where raw image-camera records are numerous and ubiquitous edge devices provide abundant distributed computing resources that can be leveraged for training.}

Despite the merits, existing distributed/federated 3D-GS training approaches still exhibit critical limitations that hinder their practical deployment. First, existing works fail to consider the resource limitations at wireless edge devices. In particular, limited GPU memory, computation, and communication capabilities of devices make it challenging to train and frequently upload large 3D‑GS models, but existing design approaches lack mechanisms to adapt model size and training workload accordingly. Second, effective aggregation of large-scale 3D-GS models is a challenging task. As the locally trained models may differ significantly in 3D structure and size, directly merging them at the central server can result in severe visual artifacts such as blurring or ghosting. Third, prior works heavily rely on a central server or network node to coordinate the devices for collecting the multi-view data and for global model refinement. However, this design deviates from practical edge scenarios, in which distributed devices need to acquire data independently based on local observations. Meanwhile, the central refinement leads to underutilization of local computational resources and raises significant privacy concerns as it often requires access to sensitive visual data (e.g., camera parameters) of local devices.

{In this paper, we develop a resource-efficient federated 3D-GS learning framework for large-scale scene reconstruction at wireless edge, where edge devices are constrained by computation, GPU memory, and communication resources. In this framework, each edge device in the wireless network performs local training based on its independently collected multi-view data and periodically uploads the model updates to the edge server without sharing raw images or camera parameters. To accommodate the constrained GPU memory, computation, and communication resources at devices for 3D-GS training, the framework incorporates a latency- and memory-aware model lightweighting mechanism and an on-server structure-consistent model recovery strategy, enabling efficient local training, privacy preservation, and effective global aggregation under constrained resources.}

\begin{itemize}
    \item In the latency- and memory-aware model lightweighting mechanism, we quantitatively evaluate the importance of different Gaussian points for each device to facilitate the pruning, and propose analytic memory consumption and end-to-end training latency models as functions of the number of Gaussian points. Then, we exploit a novel importance-to-latency ratio criterion to determine the optimal number of retained Gaussian points under GPU memory and computation/communication latency constraints, thereby maximizing the training efficiency without compromising the rendering quality.
    
    \item Furthermore, we propose a structure-consistent recovery mechanism to handle the structural mismatch among local 3D-GS models. This mismatch occurs because different devices select different subsets of Gaussian points and update them independently, leading to divergent local-model point positions and distributions. To support consistent aggregation, we introduce a binary mask to record the selected points for each device. These masks are then uploaded to the server to facilitate the recovery of complete models and gradient fields. This strategy allows the pruned Gaussian points to continue participating in future optimization, rather than being permanently excluded after pruning at any local device, ensuring a fairer contribution of each point to the global model. It preserves visual quality and mitigates visual artifacts such as ghosting and floaters.
    \item Finally, we conduct comprehensive experiments on large-scale outdoor datasets to validate the effectiveness of our framework. The results show that our method achieves faster convergence, lower training latency, and higher rendering fidelity compared with state-of-the-art federated 3D-GS benchmark schemes under constrained computation, communication, and memory budgets.
\end{itemize}

The remainder of this paper is organized as follows. Section II introduces the preliminaries of 3D-GS for scene reconstruction. Section III introduces the system model and the federated 3D-GS learning pipeline. Section IV presents the latency- and memory-aware model lightweighting mechanism. Section V proposes the model recovery strategy. Section VI presents numerical results to show the effectiveness of the proposed framework. Section VII concludes this paper.

\vspace{-0.2cm}

\section{Preliminaries}
This section provides a brief overview of the 3D-GS representation, including its model representation, splatting and rendering processes, training loss formulation, and Gaussian point density control, which serve as the foundations of our framework.

\vspace{-0.2cm}
\subsection{3D-GS Model Representation} \label{Pre_3DGS}
3D-GS is a point-based scene representation method that models a 3D scene as a collection of Gaussian points with attributes such as position, scale, rotation, opacity, and color. Its key principle is to render novel views by projecting these 3D Gaussian points onto the 2D image plane and blending their contributions. 

Specifically, each 3D Gaussian point is characterized by five attributes, including 3D position \(\boldsymbol{x} = (x, y, z) \in \mathbb{R}^{3}\), scaling factor \(\boldsymbol{s} \in \mathbb{R}^3\), rotation factor \(\boldsymbol{r} \in \mathbb{R}^4\), opacity \(\alpha \in \mathbb{R}\) with $\quad 0 \le \alpha \le 1$, and color information encoded using spherical harmonics coefficients \(\boldsymbol{h} \in \mathbb{R}^{3(v+1)^2}\), where \(v\) denotes the degree of the spherical harmonics basis. In this case, a single Gaussian point is denoted by \(\boldsymbol{g} = (\boldsymbol{x}, \boldsymbol{s}, \boldsymbol{r}, \alpha, \boldsymbol{h})\). A 3D scene is represented by a set of Gaussian points, and the corresponding 3D-GS model is given by \(\mathcal{G} = \{\boldsymbol{g}_1, \ldots, \boldsymbol{g}_U\}\), where \(U\) denotes the total number of Gaussian points. 

\vspace{-0.2cm}

\subsection{3D-GS Workflow and Optimization}
To render a 2D image from a specific viewpoint, the camera parameters are needed to specify the projection configuration. Let \(\boldsymbol{K} \in \mathbb{R}^{3 \times 3}\) and \(\boldsymbol{E} \in \mathbb{R}^{4 \times 4}\) denote the intrinsic and extrinsic matrices of the camera, respectively \cite{hartley2004multiple}. The camera configuration is then compactly represented as \(\boldsymbol{\psi} = (\boldsymbol{K}, \boldsymbol{E})\).

\subsubsection{Splatting} \label{test}

In the splatting stage, each 3D Gaussian point is projected onto the 2D image plane. The resulting 2D covariance matrix \(\boldsymbol{\Sigma}'\) is computed as $\boldsymbol{\Sigma}^{\prime}=\boldsymbol{J} \boldsymbol{W} \boldsymbol{\Sigma} \boldsymbol{W}^{\top} \boldsymbol{J}^{\top}$, where \(\boldsymbol{J} \in \mathbb{R}^{3 \times 3}\) is the Jacobian matrix obtained from an affine approximation of the projective transformation \cite{1021576}, and \(\boldsymbol{W} \in \mathbb{R}^{3 \times 3}\) corresponds to the rotation component of the camera extrinsic matrix. The 3D covariance \(\boldsymbol{\Sigma}\) is constructed by \(\boldsymbol{\Sigma} = \boldsymbol{R} \boldsymbol{S} \boldsymbol{S}^\top \boldsymbol{R}^\top\), where \(\boldsymbol{R}\) is obtained from rotation factor \(\boldsymbol{r}\) using the standard quaternion-to-matrix conversion \cite{rotation}, and \(\boldsymbol{S}\) is a positive diagonal matrix given by $\boldsymbol{S}=\text{diag}(\text{exp}(\boldsymbol{s}_x),\text{exp}(\boldsymbol{s}_y),\text{exp}(\boldsymbol{s}_z))$. Here, $\boldsymbol{s}_x, \boldsymbol{s}_y, \boldsymbol{s}_z$ are the three components of the scaling vector $\boldsymbol{s}$, $\text{exp}\left(\cdot\right)$ denotes the element-wise exponential function, and $\text{diag}\left(\cdot\right)$ constructs a diagonal matrix from its arguments.

\subsubsection{Rendering} \label{rendering}

In the rendering stage, the color of each pixel \(\boldsymbol{p} \in \mathbb{R}^2\) is computed by blending the contributions of nearby Gaussian points. Let \(\mathcal{J}\) denote the set of Gaussian points whose 2D projections overlap with \(\boldsymbol{p}\). The effective opacity \(\tilde{\alpha}_i\) of the \(i\)-th Gaussian is defined by $\tilde{\alpha}_i = \alpha_i \cdot \text{exp} \left(-\frac{1}{2} (\boldsymbol{p} - \tilde{\boldsymbol{x}}_i)^{\top} \boldsymbol{\Sigma}_i^{\prime -1} (\boldsymbol{p} - \tilde{\boldsymbol{x}}_i) \right),~i \in \mathcal{J}$, where \(\alpha_i\) is the opacity, and \(\tilde{\boldsymbol{x}}_i\) is the 2D projection of the Gaussian's center. The final color \(\boldsymbol{C}(\boldsymbol{p})\) is obtained via \(\alpha\)-blending:

\vspace{-0.5cm}

\begin{align} \label{blending}
    \boldsymbol{C}(\boldsymbol{p}) = \sum_{i \in \mathcal{J}} \boldsymbol{c}_i \tilde{\alpha}_i \prod_{j=1}^{i-1} (1 - \tilde{\alpha}_j),
\end{align}
where \(\boldsymbol{c}_i \in \mathbb{R}^3\) corresponds to the Red-Green-Blue (RGB) color space of the \(i\)-th Gaussian, computed using spherical harmonics coefficients \(\boldsymbol{h}_i\) \cite{SHreference}.

\subsubsection{Training Loss Formulation}

To optimize the model, 3D-GS employs a loss function that combines photometric accuracy and structural similarity. { Specifically, it uses the \(\mathcal{L}_1\) loss to measure the pixel-wise color error and a structural dissimilarity loss transformed from the structural similarity index measure (SSIM) \cite{SSIM} to capture perceptual quality. Following the standard 3D-GS training objective \cite{3DGS}, this structural dissimilarity loss is defined as \(\mathcal{L}_{\text{D-SSIM}} = 1 - \text{SSIM}\). Therefore, the overall loss is defined as}
\begin{align} \label{lossfunc}
    {\mathcal{L}(\boldsymbol{I}, \tilde{\boldsymbol{I}}) = (1 - \lambda)\, \mathcal{L}_1(\boldsymbol{I}, \tilde{\boldsymbol{I}}) + \lambda\, \mathcal{L}_{\text{D-SSIM}}(\boldsymbol{I}, \tilde{\boldsymbol{I}}),}
\end{align}
{ where \(\boldsymbol{I}\) is the ground-truth image, \(\tilde{\boldsymbol{I}}\) is the rendered image, and \(\lambda \in [0,1]\) is a given parameter balancing the two loss terms.} Model parameters can be optimized via gradient-based methods such as stochastic gradient descent (SGD)~\cite{3DGS}.

\subsubsection{Point Density Control} \label{densitycontrol}
3D-GS dynamically applies gradient-based densification and pruning during training to improve scene fidelity \cite{3DGS}. 

\textbf{Densification:} This process increases point density in under-represented regions by duplicating or splitting Gaussians with large reconstruction gradients.
\begin{align}
\begin{cases} 
\text{Clone}, & \text{if } \left\| \frac{\partial \mathcal{L}}{\partial \boldsymbol{x}} \right\|_2 \geq \tau_p \text{ and } \|\boldsymbol{S}\|_2 \leq \tau_s, \\
\text{Split}, & \text{if } \left\| \frac{\partial \mathcal{L}}{\partial \boldsymbol{x}} \right\|_2 \geq \tau_p \text{ and } \|\boldsymbol{S}\|_2 \geq \tau_s,
\end{cases}
\end{align}
where \(\frac{\partial \mathcal{L}}{\partial \boldsymbol{x}}\) is the gradient of the loss with respect to (w.r.t.) the 3D position, \(\boldsymbol{S}\) is the scaling matrix, and \(\tau_p\) and \(\tau_s\) are pre-defined thresholds.

\textbf{Pruning:} This removes low-contributing Gaussian points whose opacity falls below a threshold $\tau_{\text{opa}}$ (i.e., \(\alpha < \tau_{\text{opa}}\)), or whose scale is large than threshold $\tau_s$ (i.e., \(\|\boldsymbol{S}\|_2 \gg \tau_s\)).

\section{System Design and Overall Pipeline of Federated 3D-GS Learning}
This section introduces the system model and the overall learning pipeline for federated 3D-GS model training.

\vspace{-0.3cm}

\subsection{System Model}

\begin{figure*}[htbp]
    \centering
    \includegraphics[width=1\linewidth]{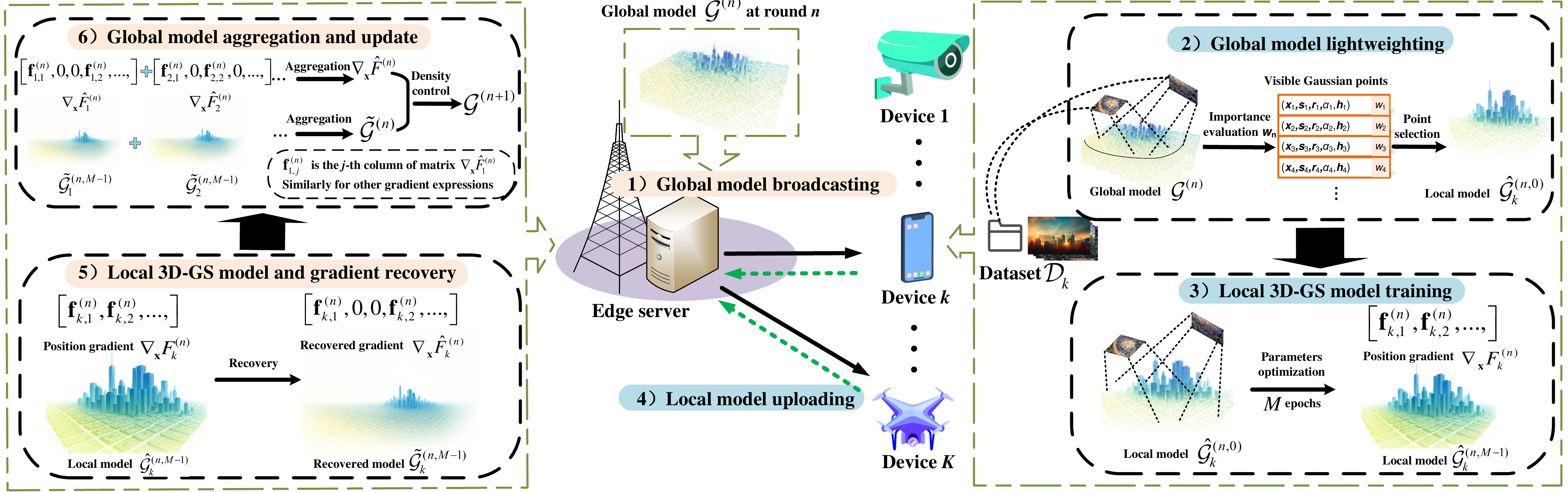}
    \caption{Illustration of the proposed federated 3D-GS learning framework. { At each communication round, the edge server broadcasts the global 3D-GS model to distributed devices; each device then performs importance-to-latency ratio-based point selection, local lightweighted training, and uploads the updated local model, position gradients, and binary mask to the server for recovery and aggregation.}
}
    \label{pipeline}
\vspace{-1.5em}
\end{figure*}

We consider a federated 3D-GS system composed of a single edge server and a set of $K >1$ distributed edge devices, denoted by $\mathcal{K} = \{1, \ldots, K\}$, as illustrated in Fig.~\ref{pipeline}. The $K$ edge devices aim to collaboratively reconstruct a large-scale 3D scene with the coordination of the edge server, under their constrained communication and computation resources.

Specifically, each device $k \in \mathcal{K}$ maintains a local 3D-GS model denoted by $\hat{\mathcal{G}}_k$, whereas the edge server aggregates and maintains a global model $\mathcal{G}$. Device $k$ independently collects a local multi-view image dataset $\mathcal{D}_k = {(\boldsymbol{I}_{k,i}, \boldsymbol{\psi}_{k,i})}_{i=1}^{D_k}$, where $\boldsymbol{I}_{k,i} \in \mathbb{R}^{L \times W \times 3}$ represents the $i$-th RGB image of resolution $L \times W$ and three color channels, and $\boldsymbol{\psi}_{k,i} \in \mathbb{R}^{11}$ denotes the corresponding camera parameters. { We assume that each device can obtain the camera extrinsic matrices in \(\{\boldsymbol{\psi}_{k,i}\}\) with respect to a common global coordinate system.\footnote{{This assumption follows related federated 3D works such as FedNeRF~\cite{fednerf} and Fed3D-GS~\cite{fed3Dgs}. In practice, a device may first establish a local coordinate system for its observed region and then register it to the global coordinate system through device-side alignment, e.g., global navigation satellite system (GNSS)-based positioning~\cite{li2023ppprtkins}, anchor-aided visual-inertial localization~\cite{xu2022omniswarm}, map-based relocalization~\cite{orbslam3}, or landmark alignment~\cite{umeyama1991}. In online scenarios, newly captured images are registered by the same device-side module before being appended to the local dataset.}}} The dataset size is $D_k = |\mathcal{D}_k|$. Note that data owned by different devices can be partially overlapping, i.e., $\mathcal{D}_k \cap \mathcal{D}_{k'} \neq \emptyset$ for some $k, k' \in \mathcal{K}$ with $k' \ne k$, where $\emptyset$ denotes the empty set. In this case, the training objective at each device $k$ is to minimize the local rendering error between the ground-truth images and the images rendered from the current 3D-GS model. Accordingly, the local loss function for device $k$ is defined as
\begin{align}
    F_k(\mathcal{G}) \triangleq \frac{1}{D_k}\sum_{( \boldsymbol{I}_{k,i},\boldsymbol{\psi}_{k,i}) \in \mathcal{D}_k} f(\mathcal{G}, \boldsymbol{I}_{k,i},\boldsymbol{\psi}_{k,i}),\quad \forall k \in \mathcal{K},
\end{align}
where $f(\mathcal{G}, \boldsymbol{I}_{k,i},\boldsymbol{\psi}_{k,i})$ denotes the per-sample loss, computed by rendering an image from model $\mathcal{G}$ using the camera parameters $\boldsymbol{\psi}_{k,i}$, and comparing it with the ground-truth image $\boldsymbol{I}_{k,i}$. The loss formulation follows the one defined in \eqref{lossfunc}. Accordingly, the global loss function on all the distributed datasets among devices is given by 
\begin{align} \label{global_loss}
    F(\mathcal{G})=\frac{1}{\sum_{k\in \mathcal{K}} D_k} \sum_{k\in \mathcal{K}} D_k F_k(\mathcal{G}).
\end{align}
The objective of federated 3D-GS learning is to minimize the global loss function in \eqref{global_loss} to obtain an optimized 3D-GS model as
\begin{align} \label{FLobjective}
    \mathcal{G}^{*}= \text{arg}~\underset{\mathcal{G}}{\text{min}}  ~F(\mathcal{G}).
\end{align}

\vspace{-0.4cm}

\subsection{Learning Pipeline} \label{learning_pipeline}
We consider the federated learning protocol based on the federated averaging (FedAvg) algorithm~\cite{fedavg}. The training process proceeds over $N$ communication rounds between the edge server and the edge devices, with each device performing $M$ local training epochs per communication round. We use the index sets $\mathcal{N} = \{0, \ldots, N-1\}$ and $\mathcal{M} = \{0, \ldots, M-1 \}$ to denote those of the communication rounds and local epochs, respectively. In this pipeline, we incorporate model lightweighting and recovery procedures, whose detailed designs are presented in Sections IV and V, respectively. The step-by-step pipeline is summarized as follows.

\subsubsection{Global Model Broadcasting}

At the beginning of each round $n \in \mathcal{N}$, the edge server broadcasts a global 3D-GS model $\mathcal{G}^{(n)} = \{ \boldsymbol{g}^{(n)}_{i} \}_{i=1}^{\bar{U}^{(n)}}$ to all devices, where $\bar{U}^{(n)}$ denotes the total number of Gaussian points. Each $\boldsymbol{g}^{(n)}_{i}$ contains the attributes of the $i$-th Gaussian point. This model contains a complete set of points representing the complete 3D scene.

\subsubsection{On-Device Model Lightweighting}
Directly training the full-size global model on edge devices is often impractical due to their limited GPU memory, high computational latency, and communication constraints. To tackle this, each device $k$ applies a local lightweighting procedure to construct a reduced model $\hat{\mathcal{G}}_k^{(n,0)}$ where $0$ denotes initial lightweighted state by selecting a subset of Gaussian points from ${\mathcal{G}}^{(n)}$ . In addition, each device $k$ generates a binary mask $\boldsymbol{m}_k^{(n)} \in \{0,1\}^{\bar{U}^{(n)}}$, where each element $m_{j}=1$ indicates that the $j$-th Gaussian point is retained, and $m_{j}=0$ indicates that it is pruned. The lightweighting operation is defined as 
\begin{align}
    (\hat{\mathcal{G}}_k^{(n,0)}, \boldsymbol{m}_k^{(n)})=\mathcal{F}(\mathcal{G}^{(n)},\{\boldsymbol{\psi}_{k,i}\}_{i=1}^{D_k}),
\end{align}
where $\mathcal{F}(\cdot)$ denotes the lightweighting function. Its design is detailed in Section~\ref{GaussianSelection}.

\subsubsection{Local Model Training}
Let $\hat{\mathcal{G}}_k^{(n,m)} $ denote the local model on device $k$ in epoch $m$ at round $n$, with the initial lightweighted model $\hat{\mathcal{G}}_k^{(n,0)}$. Each device $k$ performs local training over $M$ epochs using its lightweighted model $\hat{\mathcal{G}}_k^{(n,0)}$ and dataset $\mathcal{D}_k$. At each epoch, an image and its corresponding camera parameter are sampled and processed,\footnote{Following the original 3D-GS training framework \cite{3DGS}, at each epoch only one data sample (i.e., an image and its corresponding camera parameters) is processed due to the GPU memory limitations. Consequently, the default batch size is set to 1. Note that the design can be extended to other batch size to improve training efficiency (e.g., \cite{zhao2024scaling3dgaussiansplatting}), which is left for future work.} and thus the local gradient is computed by
\begin{align} \label{local training gradient}
        \nabla F_k^{(n, m)} = \nabla f\left(\hat{\mathcal{G}}_k^{(n,m)}, \boldsymbol{I}^{j},\boldsymbol{\psi}^j\right),
\end{align}
where $(\boldsymbol{I}^{j},\boldsymbol{\psi}^j)\sim\mathcal{D}_k$. The model is then updated by
\begin{align}
    \hat{\mathcal{G}}_k^{(n, m+1)} = \hat{\mathcal{G}}_k^{(n, m)} - \eta \nabla F_k^{(n, m)},
\end{align}
where $\eta$ is the learning rate. Meanwhile, each device accumulates the gradients w.r.t. the 3D positions of retained Gaussians across all epochs:\vspace{-0.25cm}
\begin{align}
    \nabla_{\boldsymbol{x}}{F}_k^{(n)} = \sum_{m=0}^{M-1} \nabla_{\boldsymbol{x}} F_k^{(n, m)}.
\end{align}
These spatial gradients reflect the intensity of local geometric updates and are later used by the server to perform point density control. {Note that these spatial gradient parameters do not contain raw visual data such as RGB images, viewpoint trajectories, or calibrated camera metadata, so the design reduces the direct exposure of privacy-sensitive raw records at the server.}\footnote{{Since 3D-GS is an explicit scene representation, the uploaded parameters may still encode scene geometry and appearance. Hence, the privacy benefit here is relative rather than absolute, and does not mean that the scene content is completely hidden from the server. Achieving more complete protection against information leakage from the explicit 3D-GS representation is beyond the scope of this work and is left as a promising direction for future research.}}

\subsubsection{Local Update Uploading}
Upon completing local training, each device uploads three items to the server, including the updated lightweighted model $\hat{\mathcal{G}}_k^{(n,M-1)}$, the accumulated position gradients $\nabla_{\boldsymbol{x}}{F}_k^{(n)}$, and the binary mask $\boldsymbol{m}_k^{(n)}$.

\subsubsection{On-Server Model Recovery}
Next, the edge server implements a model recovery process on the uploaded local 3D-GS models before aggregating them. A key challenge in federated 3D-GS learning lies in the inconsistent model structures across devices, due to heterogeneous lightweighting strategies. Unlike conventional FL where all devices share the same model architecture, in our case, each device selects a different subset of Gaussian points and independently optimizes their parameters. As a result, even shared points may undergo divergent spatial updates, making direct aggregation infeasible. To address this, the server employs a model recovery mechanism that reconstructs each device’s local models into a unified structure aligned with the global model. Specifically, the server restores both the updated local model and the corresponding position gradients based on the uploaded mask. The recovery is defined as
\begin{align}
    (\tilde{\mathcal{G}}_k^{(n,M-1)},\nabla_{\boldsymbol{x}}\hat{F}_k^{(n)}) = \mathcal{R}(\mathcal{G}^{(n)},\hat{\mathcal{G}}_k^{(n,M-1)}, \nabla_{\boldsymbol{x}}{F}_k^{(n)}, \boldsymbol{m}_k^{(n)}),
\end{align}
where $\mathcal{R}(\cdot)$ denotes the recovery function, for which the details are provided in Section~\ref{Aggregation}. 

\subsubsection{Global Model Aggregation and Update}
After recovery, the server performs weighted averaging across the recovered models and gradients, which is given by
\begin{align}
    \tilde{\mathcal{G}}^{(n)} = \sum_{k \in \mathcal{K}} \left( \frac{D_k}{\sum_{k \in \mathcal{K}} D_{k}} \tilde{\mathcal{G}}_k^{(n,M-1)} \right).
\end{align}
Similarly, the aggregated global position gradient is given by
\begin{align} \label{averaginggrad}
    \nabla_{\boldsymbol{x}}\hat{F}^{(n)}  = \sum_{k \in \mathcal{K}}\left(\left(\frac{{D}_k}{\sum_{k\in \mathcal{K}} {D}_k}\right)\nabla_{\boldsymbol{x}}\hat{F}_k^{(n)}\right).
\end{align}
Then, the server applies density control based on $\nabla_{\boldsymbol{x}}\hat{F}^{(n)}$ to either prune or densify Gaussians, yielding the updated global model $\mathcal{G}^{(n+1)}$ for the next round. The previous global model is discarded, and the new one is broadcast to all devices, for local model update and aggregation in round $n+1$.

\vspace{-0.2cm}
\section{Latency- and Memory-Aware 3D-GS Model Lightweighting} \label{model_lightweighting}

To enable efficient large-scale 3D-GS model training subject to resource constraints at edge devices, we design a model lightweighting mechanism to adaptively select and remove unnecessary Gaussian points before each round of local training. This design is motivated by the observation that only a portion of the Gaussian points significantly contribute to the rendered images due to the viewpoint diversity across local datasets, and as a result, pruning the low-contributing points helps reduce memory usage and training latency overhead without compromising the rendering quality.

To guide the Gaussian point selection strategy in the lightweighting design, we introduce a novel metric termed importance-to-latency ratio, which evaluates rendering utility relative to training latency cost. To clearly establish the quantitative relationship between the number of Gaussian points and the resource cost, we provide a concrete model on the GPU memory consumption and the latency, based on low-level GPU execution behaviors. As such, the lightweighting pipeline follows three steps: (i) estimating per-point importance from local viewpoints, (ii) modeling memory and latency overhead during training, and (iii) selecting a subset of Gaussian points that maximizes the importance-to-latency ratio under memory and latency constraints. 

\vspace{-0.3cm}

\subsection{Importance Score Estimation}
To assess the contribution of each Gaussian point to the final rendered output, we compute a device-specific importance score for every point based on local viewpoints\footnote{{Here, local viewpoints refer to the camera parameters of the images already available in the local dataset at device \(k\). If new images are later collected from different viewpoints, they can be incorporated into the local dataset, and the importance scores and selected Gaussian subset can be recomputed in subsequent communication rounds. For novel rendering within the observed scene, the aggregated 3D-GS model can render unseen viewpoints, while completely unobserved regions require additional visual observations and subsequent federated updates.}}. The core idea is that the points frequently observed and actively contributing to the rendering quality from a device's camera views should be prioritized during training, while those with minimal visual impact or being frequently occluded can be safely removed.

Recall that, at the beginning of communication round $n$, the broadcasted global model is denoted by $\mathcal{G}^{(n)}=\{\boldsymbol{x}_{j}^{(n)}, \boldsymbol{s}_{j}^{(n)}, \boldsymbol{r}_{j}^{(n)}, \alpha_{j}^{(n)}, \boldsymbol{h}_{j}^{(n)}\}_{j=1}^{\bar{U}^{(n)}}$. Each device $k$ has its local dataset with camera parameters $(\boldsymbol{\psi}_{k,i})_{i=1}^{D_k}$, where each $\boldsymbol{\psi}_{k,i}$ defines an imaging plane. A Gaussian point is said to intersect a view if it is projected onto the associated imaging plane and contributes to at least one pixel. Let $\mathcal{P}_d$ denote the set of pixels in the $d$-th imaging plane, and $\mathcal{S}_d^{(n)} \subseteq \{1,\ldots,\bar{U}^{(n)}\}$ represent the indices of Gaussian points that intersect this view. {Inspired by importance-based Gaussian evaluation \cite{eagles, lightgs, minisplat},} for a given Gaussian point $j$, its influence on the $d$-th view is quantified as\vspace{-0.2cm}
\begin{align} \label{importance}
    w_j^{d} =\left\{
	\begin{aligned}
	& \sum_{i \in \mathcal{P}_d} \alpha_{j,i}^{(n)}\prod_{\substack{l\in {\mathcal{S}}_d^{(n)} \\ \text{dis}(l) < \text{dis}(j)}}\left(1-\alpha_{l,i}^{(n)}\right) \quad && \text{if}~j \in {\mathcal{S}}_d^{(n)}, \\
	& 0 \quad && \text{if}~j \notin {\mathcal{S}}_d^{(n)},\\
	\end{aligned}
    \right.
\end{align}
where $\alpha_{j,i}^{(n)}$ denotes the opacity contribution of point $j$ to pixel $i$, and $\text{dis}(j)$ is its depth relative to that pixel. The product term accounts for occlusion by front-facing points, i.e., only unoccluded contributions are aggregated. In essence, $w_j^d$ measures the impact of point $j$ on the $d$-th image after occlusion handling. If the point does not project onto this view, its contribution is set to zero.

{In our work, the importance score is computed from the same forward rasterization quantities used by 3D-GS training. On the splatting side, the score requires the projected Gaussian footprint, intersected pixels, and depth ordering, which are also used in the splatting stage. On the rendering side, it uses the opacity contribution and front-to-back transmittance product in the standard \(\alpha\)-blending process. Thus, importance estimation can reuse the same splatting/rendering pipeline and, when the model state and view are unchanged, the corresponding intermediate values/caches. The additional operation is mainly the accumulation of per-pixel contributions into a per-Gaussian score vector, without an additional backward pass, optimizer update, or communication step. In our latency accounting, this forward-only pre-lightweighting cost is covered by the same splatting/rendering latency model and is much smaller than the repeated forward-backward local training and model synchronization costs.}

Next, we define the overall importance score of point $j$ across the entire local dataset as the sum of its contributions over all camera views:
\begin{align}
    w_j = \sum_{d=1}^{D_k} w_j^{d}.
\end{align}
As a result, we obtain a per-device importance score $\boldsymbol{w}_k \in \mathbb{R}^{\bar{U}^{(n)}}$ for device $k$, where the $j$-th entry $w_j$ represents the importance score for the $j$-th point. 

\begin{figure}[t]
  \centering
  \includegraphics[width=0.55\linewidth]{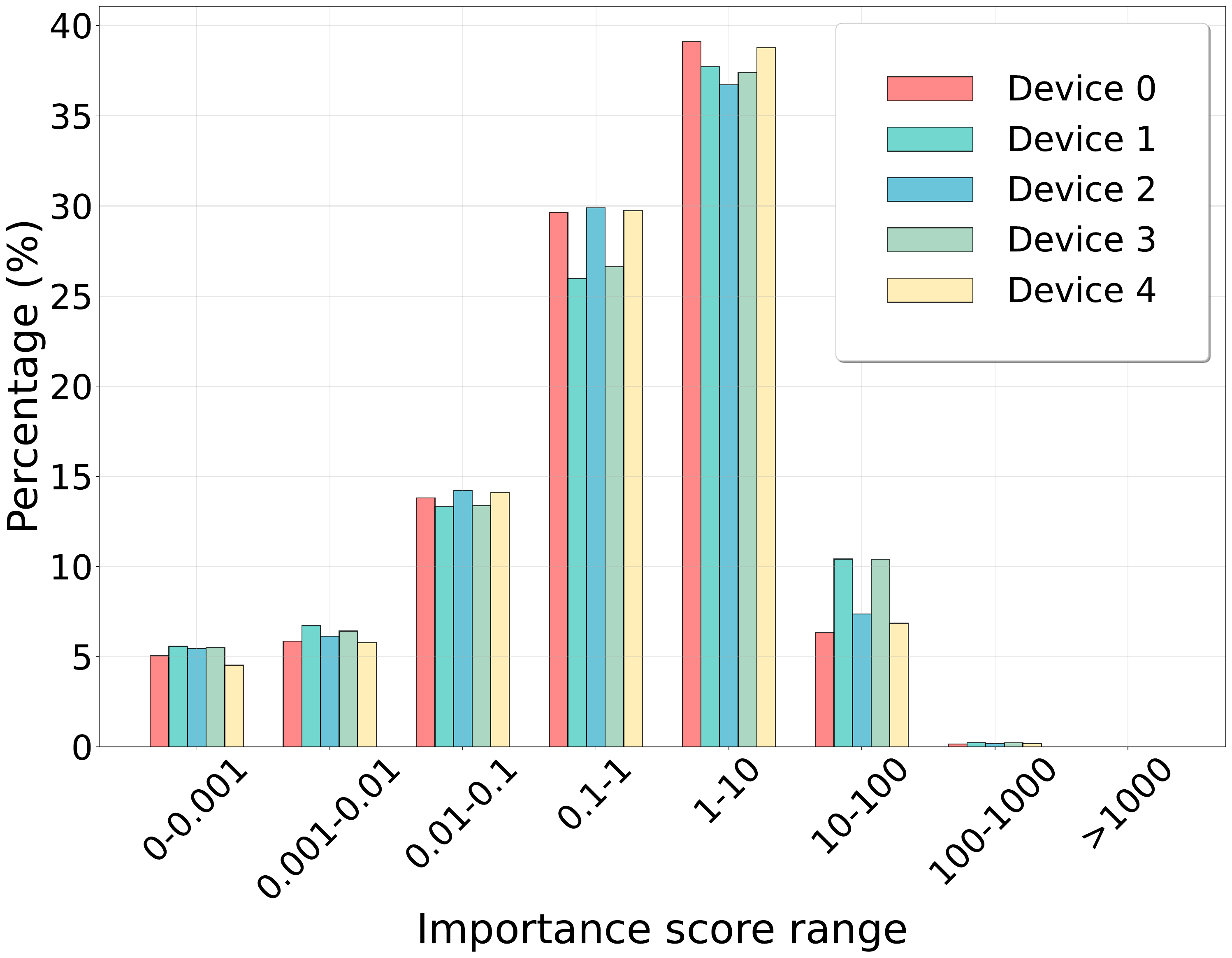}  
  \caption{Distribution and range of importance scores for Gaussian points in an example from experiments with 5 devices and a total of 3,222,794 Gaussian points.}
  \label{Importance_score}
\vspace{-1.8em}
\end{figure}

{Intuitively, this score quantifies the visible rendering contribution of a Gaussian point from the local viewpoints available to device $k$. A high score indicates that the point is frequently visible, contributes to many pixels with non-negligible opacity, and is not heavily occluded, whereas a low score indicates that the point is rarely observed, weakly contributes to the rendered pixels, or is frequently occluded. Besides, the absolute score values may vary with the density of local camera views, device deployment, and Gaussian points. First, increasing the density of local camera views at device $k$ adds more view-wise terms to $w_j=\sum_{d=1}^{D_k}w_j^d$, so points that are repeatedly visible from these views tend to receive larger scores. Second, since the score is computed locally, increasing the device density does not change the score at any single device; instead, a point that is invisible or occluded at one device may still be visible and receive a high score at another, so scores from different devices reflect their own local viewpoints and are not directly comparable across devices. Third, when the same scene region is represented by more Gaussians, its nearly conserved per-pixel rendering contribution is shared among more primitives, so the score of each individual Gaussian tends to decrease. In all these cases, however, only the numerical values of the scores change, whereas the meaning and role of the importance score stay the same: it always measures how much each Gaussian point contributes to rendering the local views of its device, and is used to rank points within the same global model. For this reason, the score is meaningful as a relative ranking on each device, rather than as an absolute value that can be compared across different models or scenes.} These scores are then used to guide point selection under memory and latency constraints. An example of the computed importance score distribution across devices is illustrated in Fig.~\ref{Importance_score}. It is observed that the majority of points have extremely low importance scores, implying their minimal influence on the rendering result. This demonstrates that a substantial portion of the points is visually redundant and can be removed without affecting rendering fidelity.

\vspace{-0.25cm}
\subsection{GPU Memory and Latency Modeling}

In this subsection, we model the GPU memory and end-to-end latency cost of 3D-GS training as explicit functions w.r.t the number of selected Gaussian points by considering low-level GPU execution behaviors, establishing a clear quantitative relationship between the number of retained points and the associated resource overhead.

\subsubsection{GPU Memory Consumption Modeling} \label{memory model}
We begin with a memory consumption model that characterizes the peak GPU usage during local training. This model is used to avoid out-of-memory failures and ensure safe execution on edge devices. Based on the memory consumption model in \cite{kim2024llmemestimatinggpumemory, 3417050}, we decompose the peak memory usage during the $n$-th communication round on device $k$ into four major components, including base memory consumption, as well as usage for model parameters, optimizer states, and input/output tensors, respectively.

First, we consider the base memory consumption $O_k^{\text{base}}$, which accounts for the memory occupied by the GPU runtime system, such as CUDA context and kernel management \cite{9940581}. This component is independent of the model or data and remains constant throughout training.

Then, the memory consumption from model parameters is given by
\vspace{-0.15cm}
\begin{align}
    O_k^{(n), \text{para}}( U_k^{(n)})=z_{\text{para}}q_{\text{para}} U_k^{(n)},
\end{align}
where $U_k^{(n)}$ is the number of selected Gaussian points, $z_{\text{para}}$ is the number of parameters per point, and $q_{\text{para}}$ is the number of Bytes per parameter (e.g., 4 Bytes for FP32).

Next, we consider the memory consumption from the optimizer. In this work, we consider the SGD optimizer. Accordingly, the memory consumption by  optimizer is
\begin{align}
    O_k^{(n),\text{opt}}( U_k^{(n)}) = z_{\text{opt}}q_{\text{opt}} U_k^{(n)},
\end{align}
where $ z_{\text{opt}} = z_{\text{para}} $ denotes the required number of gradients for one single Gaussian point, that equals to the number of parameters, and $q_{\text{opt}}$ is the required Bytes per gradient entry.

Finally, we analyze the memory consumption of input and output tensors during training, including input data samples,\footnote{{Following the official 3D-GS implementation \cite{3DGS}, the entire local dataset is preloaded into GPU memory at the beginning of training and kept resident throughout, so as to avoid frequent host-to-GPU data transfers at each iteration.}} intermediate outputs, and predicted data samples \cite{3417050}. The memory consumption of the tensors is given by \vspace{-0.13cm}
\begin{align}
    O_k^{n, \text{ten}}( U_k^{(n)}) = D_k(z_{\boldsymbol{\psi}}q_{\text{para}}+3|{\boldsymbol{I}}|q_{\text{para}}) +z_{\text{temp}}q_{\text{para}} U_k^{(n)},
\end{align}
where $z_{\boldsymbol{\psi}}$ denotes the number of elements in the camera parameter matrix, which serve as input, $|{\boldsymbol{I}}|$ is the number of pixels in the ground-truth image and $z_{\text{temp}}$ is the number of the elements of the temporary intermediate variables (e.g., covariance matrices) used during rendering.

Combining the above components, the peak memory model of device $k$ at the $n$-th communication round is expressed by \vspace{-0.1cm}
\begin{align}  
    O_k^{(n), \text{peak}} ( U_k^{(n)}) &= O_k^{(n), \text{para}}+ O_k^{(n), \text{opt}}+O_k^{(n), \text{ten}} + O_k^{ \text{base}} \nonumber \\ & =  (z_{\text{para}}q_{\text{para}}+z_{\text{opt}}q_{\text{opt}}+z_{\text{temp}}q_{\text{para}} )  U_k^{(n)} \nonumber \\ &\quad\quad+D_k(z_{\boldsymbol{\psi}}q_{\text{para}}+3|{\boldsymbol{I}}|q_{\text{para}}) + O_k^{\text{base}}. \label{eq:peakmem}
\end{align}
To prevent out-of-memory issues, we enforce the following constraint:
\vspace{-0.13cm}
\begin{align}
O_k^{(n), \text{peak}} ( U_k^{(n)})  \leq O_k^{\text{max}}, \label{eq:memconstraint}
\end{align}
where $O_k^{\text{max}}$ denotes the maximum GPU memory of device $k$.

{We compare the peak-memory model in \eqref{eq:peakmem} with the measured peak GPU memory. As reported in Table~\ref{tab:mem_validation}, the prediction matches the measurement within about $1.2\%$ (mean absolute percentage error $1.0\%$), showing that our memory model provides an accurate estimate with an acceptable error.}

\begin{table}[t]
\centering
\caption{Validation of the Analytical Peak-Memory Model in \eqref{eq:peakmem} Against the Measured Peak GPU Memory During Local 3D-GS Training.}
\label{tab:mem_validation}
\renewcommand{\arraystretch}{1.2}
\begin{tabular}{ccccc}
\toprule
$U_k$ ($\times 10^6$) & $D_k$ & Predicted [GB] & Measured [GB] & Rel. err. [\%] \\
\midrule
0.756472 & 320 & 5.309 & 5.265 & 0.84 \\
0.934897 & 320 & 5.392 & 5.329 & 1.18 \\
1.132375 & 320 & 5.484 & 5.426 & 1.07 \\
\bottomrule
\end{tabular}
\end{table}

\subsubsection{End-to-End Latency Modeling}

We next model the end-to-end latency of federated 3D-GS training, which consists of computation latency and transmission latency.\footnote{We do not include the latency of the sparse point cloud initialization (e.g., via COLMAP \cite{sfm}), as we focus on analyzing the latency during training. Efficient and federated-friendly initialization methods are considered out of the scope of this work and can be explored in future research.}

First, we consider the computation latency, which stems from the execution of GPU kernels during local training. Unlike conventional latency estimations that rely solely on floating-point operation counts (FLOPs), our analysis captures the unique data access patterns and parallel execution characteristics of 3D-GS with low-level GPU behaviors. Specifically, the computation latency is determined by the two stages of the 3D-GS framework: splatting and rendering. 

In the splatting stage, as presented in Section \ref{test}, each Gaussian point undergoes 3D-to-2D projection with covariance transformation. This is implemented using a point-parallel execution pattern, where each GPU thread independently processes a single Gaussian point across both forward and backward processes during training. Based on the GPU execution model in \cite{GPURUNINGTIME}, the latency for the splatting stage for device $k$ in epoch $m$ at communication round $n$ is given by
\begin{align} \label{splat_latency}
    T_{k,\text{splat}}^{(n,m)} (U_k^{(n)})= \left\lceil {\frac{U_k^{(n)}}{{\beta_{k}}}} \right\rceil \gamma_k,
\end{align}
where $\beta_k$ is the maximum number of GPU threads that can be run in parallel, and $\gamma_k$ is a device-specific latency coefficient that encapsulates the arithmetic cost and throughput of processing one block of Gaussian points. This formulation captures the linear growth of latency with the number of points in the 3D-GS model. \vspace{-0.08cm}
\begin{remark}
    The coefficient $\gamma_k$ in (\ref{splat_latency}) is derived from the GPU execution and roofline performance model~\cite{GPURUNINGTIME},  given by \vspace{-0.1cm}
    \begin{align}
        \gamma_k = \frac{N_{\text{FLOPs}}}{\epsilon_k N_k^{\text{roofline}} },~N_k^{\text{roofline}} = \min \left( \frac{N_{\text{FLOPs}}}{B}  B_k, \varphi_{k} \right),
    \end{align}
    where $N_{\text{FLOPs}}$ denotes the FLOPs for a single thread block (e.g., 256 Gaussian points), $\epsilon_k$ is the effective computational efficiency, $B$ is the volume of access data per block, $B_k$ and $\varphi_k$ denote the maximum memory bandwidth and the peak FLOP throughput of device $k$, respectively. This formulation captures both the computational demands of the 3D-GS algorithm and the hardware capabilities of the GPU. Since all the involved parameters can be profiled during training, the model is readily applicable for estimating in practice.
\end{remark}

In the rendering stage, as presented in Section \ref{rendering}, each image pixel is computed by accumulating the contributions of overlapping Gaussian points using the $\alpha$-blending technique \cite{3DGS}. This process adopts a different pattern: each GPU thread computes the color of a single pixel in parallel. Accordingly, the latency of the rendering stage is computed by 
\begin{align} \label{render_latency}
    T_{k,\text{render}}^{(n,m)} (U_k^{(n)})=  \left\lceil {\frac{|{\boldsymbol{I}}|}{{{\beta}_{k}}}} \right\rceil \delta(U_k^{(n)})\tilde{\gamma}_k,
\end{align}
where $\tilde{\gamma}_k=\frac{\tilde{N}_{\text{FLOPs}}}{\tilde{\epsilon}_k\tilde{N}_{k}^{\text{roofline}}}$ denotes the latency coefficient for the rendering stage and $\tilde{\epsilon}_k$ is the efficiency during rendering. The term $\tilde{N}_k^{\text{roofline}} = \min \left( \frac{\tilde{N}_{\text{FLOPs}}}{\tilde{B}} B_k, {\varphi}_k \right)$ follows the same roofline model as in the splatting stage, but varies in rendering-specific FLOPs $\tilde{N}_{\text{FLOPs}}$ and memory volume $\tilde{B}$. To account for the increased computational burden when rendering denser point set in one pixel, we introduce a scaling factor $\delta(U_k^{(n)}) = \phi_1 + \phi_2 \frac{U_k^{(n)}}{|\boldsymbol{I}|}$, where $\phi_1$ and $\phi_2$ are empirically fitted constants. This term captures the linear growth in per-pixel computation as the number of visible Gaussian points increases. Therefore, the total computation latency after $M$ local epochs is calculated by $T_{k, \text{comp}}^{(n)}(U_k^{(n)})= M(T_{k,\text{splat}}^{(n,m)}(U_k^{(n)})+T_{k,\text{render}}^{(n,m)}(U_k^{(n)}))$.

\begin{table}[!t]
\centering
\caption{Estimation and Measurement of Latency for Two Stages (Splatting and Rendering) for 100 Local Epochs.}
\label{estimation_latency}
\begin{tabular}{|c|cc|cc|}
\hline

& \multicolumn{2}{c|}{$T_{\text{splat}}(\text{s})$} 
& \multicolumn{2}{c|}{$T_{\text{render}}(\text{s})$} \\
\cline{2-5}
& Measured & Estimated & Measured & Estimated \\
\hline
RTX 3090  & 1.5530 & 2.4728  & 41.3355 & 43.8591 \\
\hline
RTX 3060  & 4.1076 & 5.9211 & 72.3471 & 75.9602 \\
\hline
\end{tabular}
\vspace{-1.7em}
\end{table}

To validate the training latency models presented in \eqref{splat_latency} and \eqref{render_latency}, we conduct experiments to compare the estimated latency against the actual measurements obtained using the NVIDIA Nsight Profiler~\cite{Nsight2023} on two different GPU devices, as shown in Table \ref{estimation_latency}. The results suggest that the proposed computation latency model provides a reasonably accurate estimation of the actual 3D-GS training latency.

Next, we consider the communication latency, which includes downlink and uplink latency. The downlink latency corresponds to the time duration for broadcasting the global model $\mathcal{G}^{(n)}$ from the edge server to device $k$, which is calculated by $T_k^{(n), \mathrm{d}}= \frac{H^{(n)}}{r_k^{(n), \mathrm{d}}}$, where $H^{(n)} = z_{\text{para}} q_{\text{para}}\bar{U}^{(n)}$ is the total number of Bytes required to transmit the model, and $r_k^{(n), \mathrm{d}}$ denotes the downlink data rate. Here, $\bar{U}^{(n)}$ is the known number of Gaussian points of the global model, which is fixed within each communication round, and the communication rate is constant per round.

The uplink latency accounts for the transmission of the local model $\mathcal{G}_k^{(n, M-1)}$, the binary mask $\boldsymbol{m}_k^{(n)}$, and the accumulated position gradients $\nabla_{\boldsymbol{x}}{F}_k^{(n)}$. Since the binary mask introduces negligible overhead, it is omitted in the size calculation. The total required number of Bytes for uplink transmission are $H_k^{(n)}(U_k^{(n)})= z_{\text{para}} q_{\text{para}}U_k^{(n)} + \text{dim}(\boldsymbol{x}) q_{\text{opt}}U_k^{(n)}$, where $\text{dim}(\boldsymbol{x}) = 3$ represents the dimensionality of the 3D position parameters. The corresponding uplink latency is $T_k^{(n), \mathrm{u}}= \frac{H_k^{(n)}(U_k^{(n)})}{r_k^{(n), \mathrm{u}}}$, with $r_k^{(n), \mathrm{u}}$ denoting the uplink rate for device $k$.

By combining the above components, the total latency for device $k$ at the $n$-th communication round is given by \vspace{-0.08cm}
\begin{align} \label{latency}
    T_k^{(n)}(U_k^{(n)}) &= T_{k,\text{comp}}^{(n)}(U_k^{(n)}) + T_k^{(n), \mathrm{d}} + T_k^{(n), \mathrm{u}}(U_k^{(n)})\nonumber \\
    &= M  \left( \left\lceil \frac{U_k^{(n)}}{\beta_k} \right\rceil \frac{N_{\text{FLOPs}}}{\epsilon_k N_k^{\text{roofline}}}  \right. \nonumber \\
    &\quad \left. + \left\lceil \frac{|{\boldsymbol{I}}|}{{\beta}_{k}} \right\rceil \delta(U_k^{(n)}) \frac{\tilde{N}_{\text{FLOPs}}}{\tilde{\epsilon}_k\tilde{N}_k^{\text{roofline}}} \right)   \nonumber \\ 
    &\quad  + \frac{z_{\text{para}} q_{\text{para}}\bar{U}^{(n)}}{r_k^{(n), \mathrm{d}}}\!+\! U_k^{(n)}\frac{z_{\text{para}} q_{\text{para}} + \text{dim}(\boldsymbol{x}) q_{\text{opt}}}{r_k^{(n), \mathrm{u}}}.
\end{align}

\vspace{-0.1cm}
{ \begin{remark}
It is worth noting that the number of local epochs $M$ is another important factor affecting the end-to-end latency, as shown in \eqref{latency}. Under a fixed latency budget, there exists a tradeoff between retaining more Gaussian points for fewer local epochs and retaining fewer Gaussian points for more local epochs. The former improves the coverage of the current global 3D-GS model but provides fewer optimization steps for each retained point, whereas the latter enables more sufficient local refinement of the selected important points but may exclude some less highly ranked points that are still useful for global reconstruction. In this work, we fix $M$ across all compared methods and focus on optimizing the number of retained Gaussian points $U_k^{(n)}$, so as to isolate the effect of the proposed latency- and memory-aware lightweighting design. Jointly optimizing $M$ and $U_k^{(n)}$ would require modeling the long-term convergence behavior across communication rounds and is left as an interesting direction for future work.
\end{remark}}

\vspace{-0.2cm}

\subsection{Importance-to-Latency Ratio-Based Point Selection} \label{GaussianSelection}

We now develop a point selection strategy that leverages the per-point importance scores and the models of memory and latency cost. The objective is to select a subset of Gaussian points that provides the greatest rendering benefit while incurring the least latency overhead under the device's resource constraints. In particular, we present a novel metric, named the importance-to-latency ratio, which is designed to maximize the rendering contribution per unit of end-to-end training latency.

Let $U_k^{(n)}$ denote the number of Gaussian points selected for training on device $k$ at the $n$-th communication round. We define the importance-to-latency ratio as
\begin{align}
    \rho_k^{(n)}(U_k^{(n)})= \frac{\sum_{j=1}^{U_k^{(n)}}\tilde{w}_j}{T_k^{(n)}(U_k^{(n)})},
\end{align}
where $\{\tilde{w}\}_{j=1}^{\bar{U}^{(n)}}$ is the sequence of importance scores sorted in descending order, i.e., $\tilde{w}_1 \geq \tilde{w}_2\geq \ldots\geq\tilde{w}_{\bar{U}^{(n)}} > 0$, the term $\sum_{j=1}^{U_k^{(n)}}\tilde{w}_j$ represents the cumulative importance of the top-$U_k^{(n)}$ points, and $T_k^{(n)}(U_k^{(n)})$ denotes the corresponding end-to-end latency, as given in \eqref{latency}.

We now leverage the importance-to-latency ratio $\rho_k^{(n)}(U_k^{(n)})$ to guide resource-efficient Gaussian point selection. Intuitively, maximizing this ratio allows each device to select the subset of points that offers the highest cumulative rendering benefit per unit of training cost. This leads to the following optimization problem, which determines the optimal number of selected points under device-specific memory and latency constraints:
\begin{align} \label{maxratio}
\mathop {\max }\limits_{U_k^{(n)} \in \mathbb{Z}_+}  &\quad  \rho _k^{(n)}(U_k^{(n)})  \\
{\rm{ s.t.}}  &\quad 0 \le O_k^{(n), \text{peak}} ( U_k^{(n)})  \leq O_k^{\text{max}}, \label{GPUconstraint} \\
                & \quad 0 \le T_k^{(n)}(U_k^{(n)}) \le T_{\max}, \label{latencyconstraint}
\end{align}
where the constraint in \eqref{GPUconstraint} corresponds to the GPU memory constraint, and the constraint in \eqref{latencyconstraint} is the end-to-end training latency constraint. In \eqref{latencyconstraint}, $T_{\max}$ represents the maximum tolerable per-device latency, which is introduced to ensure that the training is completed within an acceptable duration.

\begin{remark}
{The optimization problem in \eqref{maxratio}--\eqref{latencyconstraint} is solved online by each device at each communication round, rather than as a global optimization over the entire training horizon that pre-determines the retained Gaussian configuration of all rounds. This per-round adaptation is adopted because the Gaussian importance scores, the feasible memory and latency budgets, and the wireless channel conditions are all round-dependent and not available in advance.}
\end{remark}

Due to the monotonicity of $O_k^{(n), \text{peak}} ( U_k^{(n)})$ and $T_k^{(n)}(U_k^{(n)})$ w.r.t. the number of selected points $U_k^{(n)}$, the constraints in \eqref{GPUconstraint} and \eqref{latencyconstraint} can be equivalently translated into two upper bounds on $U_k^{(n)}$. Let $\tilde{U}_k$ and $U_k^{\text{lat}}$ denote the maximum number of Gaussian points satisfying the memory and latency constraints, respectively. Both of them can be efficiently obtained by binary search. Then, the feasible range of $U_k^{(n)}$ is transformed to $1 \le U_k^{(n)} \le \hat{U}_k \triangleq \min\left\{ \tilde{U}_k,\ U_k^{\text{lat}} \right\}$.

Notice that the latency function in \eqref{latency} contains a ceiling operator, which introduces stepwise changes in latency function and disrupts the regularity of $\rho_k^{(n)}(U_k^{(n)})$ over the entire feasible domain. To tackle this issue, we partition the candidate set of Gaussian point counts into smaller intervals and boundary points. Within each interval, $\rho _k^{(n)}(U_k^{(n)})$ behaves more smoothly and can be characterized efficiently. Specifically, we define
\begin{align} \label{subset}
    \mathcal{B}_b  \triangleq  
    \{\,&U_k^{(n)} \in \mathbb Z_+ :\ \nonumber \\ &b\beta_k+1 \le U_k^{(n)} \le\min\{(b+1)\beta_k-1,\hat{U}_k\}\}, \nonumber \\ 
    &b=0,1,\ldots,\lfloor\frac{\hat{U}_k-1}{\beta_k}\rfloor,
\end{align}
\vspace{-0.25cm}and
\begin{align}
    \mathcal{B}_{\text{edge}}\!=\!\{ U_k^{(n)}\in \mathbb Z_+ \!: U_k^{(n)} \!=\!(\tilde{b}+1)\beta_k \},\nonumber \\  \tilde{b}=0, 1,\ldots,\lfloor\frac{\hat{U}_k}{\beta_k}\rfloor-1.
\end{align}
For each interval $\mathcal{B}_b$, we aim to find the locally optimal number of selected points, denoted by $U_{k,b}^{(n)*}$, which maximizes the importance-to-latency ratio.
\begin{theorem} \label{theorem}
For each set $\mathcal{B}_b$, there exists at least one optimal solution $U_{k,b}^{(n)*} \in \mathcal{B}_b$, such that
\begin{align}
    \Delta \rho_k^{(n)}(U_k^{(n)}) \geq 0, \quad \forall U_k^{(n)} \in \mathcal{B}_b \text { with } U_k^{(n)}<U_{k,b}^{(n)*},
\end{align}
and
\begin{align}
    \Delta \rho_k^{(n)}(U_k^{(n)}) \le 0, \quad \forall U_k^{(n)}+1 \in \mathcal{B}_b \text { with } U_k^{(n)} \geq U_{k,b}^{(n)*},
\end{align}
where $\Delta \rho_k^{(n)}(U) \triangleq\rho_k^{(n)}(U+1) - \rho_k^{(n)}(U)$.
\begin{proof}
 See Appendix \ref{appendix_1}.
\end{proof}
\end{theorem}
Theorem \ref{theorem} implies that within each $\mathcal{B}_b$, the ratio $\rho_k^{(n)}(U_k^{(n)})$ should be entirely non-decreasing, entirely non-increasing, or first increasing to a maximum and then decreasing. This allows us to efficiently find $U_{k,b}^{(n)*}$ via discrete binary search. After identifying the locally optimal $U_{k,b}^{(n)*}$ for each block $\mathcal{B}_b$ and evaluating the endpoints in $\mathcal{B}_{\text{edge}}$, the global optimum is given by
\begin{align}
    U_k^{(n)*} = \text{arg} \max\{\rho_k^{(n)}(U_{k,b}^{(n)*}), \rho_k^{(n)}((\tilde{b}+1)\beta_k)\},  \nonumber\\ b=0,1,\ldots,\lfloor\frac{\hat{U}_k-1}{\beta_k}\rfloor,~\tilde{b}=0, 1,\ldots,\lfloor\frac{\hat{U}_k}{\beta_k}\rfloor-1.
\end{align}

To track Gaussian points, we employ a binary mask vector $\boldsymbol{m}_k^{(n)}=[m_1,m_2,\ldots,m_{\bar{U}^{(n)}}] \in \{0,1\}^{\bar{U}^{(n)}} $, where \(m_j = 1\) indicates that the $j$-th Gaussian point is retained, and $m_j = 0$ otherwise. The number of selected points satisfies $\| \boldsymbol{m}_k^{(n)}\|_0 = U_k^{(n)}$, where $\|\cdot\|_0$ denotes the $l_0$-norm. The mask is then transmitted from each device to the edge server, facilitating global model aggregation. Notably, the use of such a compact mask offers a communication-efficient and structure-preserving approach for model pruning, which is particularly advantageous for large-scale models.

In summary, the proposed model lightweighting mechanism enables each device to only retain a subset of Gaussian points that balance the rendering quality and training latency cost. This is achieved by maximizing the importance-to-latency ratio under per-device memory and latency constraints. The optimization problem can be solved efficiently via block-wise search, and the selected subset is encoded using a binary mask to facilitate the aggregation on server.

\section{Model and Gradient Recovery for Structure-Consistent Aggregation} \label{Aggregation}

In this section, we present the model and gradient recovery mechanism to address the structural inconsistency issue across locally trained 3D-GS models caused by heterogeneous lightweighting strategies of devices. Since each device selects and updates a distinct subset of Gaussian points, the number and structure of Gaussian points vary across devices, breaking the structural alignment required for standard aggregation methods such as parameter-wise averaging. Directly merging the uploaded models would lead to severe visual artifacts (e.g., ghosting or blurring), particularly near object boundaries~\cite{Lin2024CVPR}. To address this, we propose a recovery mechanism that restores both the local model parameters and the associated position gradients to unified structures before aggregation, thereby ensuring consistent indexing and alignment across all devices.

In particular, after completing local training, device $k$ uploads the following parameters to the edge server: (i) its updated local 3D-GS model $\hat{\mathcal{G}}_k^{(n,M-1)}$, (ii) the binary mask $\boldsymbol{m}_k^{(n)}$ indicating which Gaussians were retained, and (iii) the accumulated gradients $\nabla_{\boldsymbol{x}}{F}_k^{(n)}$ for 3D positions. Denote $\boldsymbol{m}_k^{(n)} \in \{0,1\}^{\bar{U}^{(n)}}$ as the binary mask uploaded by device $k$, where $m_i =1$ indicates that the $i$-th Gaussian point was locally retained and uploaded. Let $\boldsymbol{g}_{k,j}^{(n)}$ denote the $j$-th point in the uploaded local model (after lightweighting), and $\boldsymbol{g}_i^{(n)}$ denote the $i$-th point in the global model distributed at the beginning of the current communication round. Since the retained Gaussians form a sparse subset of the global model, we define the prefix sum $j=\sum_{\ell=1}^i m_{\ell}$ to map the global index $i$ to the corresponding local index $j$ in the uploaded model. As such, the parameters of the recovered point at index $i$ for device $k$ are given by
\vspace{-0.1cm}
\begin{align}
    \tilde{\boldsymbol{g}}^{(n)}_{k,i}=\left\{\begin{array}{ll}
\hat{\boldsymbol{g}}^{(n)}_{k,j}, & \text { if } m_{i}=1, \\
\boldsymbol{g}^{(n)}_i, & \text { if } m_{i}=0.
\end{array} \right.
\end{align}
This mechanism implies:
\begin{itemize}
    \item If the point was selected during lightweighting and has been updated via local training, we reuse the corresponding entry from the uploaded model.
    \item If the point was pruned and not updated locally, we recover it by copying the original Gaussian from the global model at the same index.
\end{itemize}
Similarly, $\boldsymbol{f}^{(n)}_{k,j}$ is the accumulated gradient of the $j$-th point in the uploaded gradient. To reconstruct the complete gradient for global aggregation, the recovered gradient for the point at index $i$ is constructed by 
\begin{align}
    \hat{\boldsymbol{f}}^{(n)}_{k,i}=\left\{\begin{array}{ll}
\boldsymbol{f}^{(n)}_{k,j}, & \text { if } m_{i}=1, \\
\boldsymbol{0}, & \text { if } m_{i}=0,
\end{array} \right.
\end{align}
where $\boldsymbol{0}$ denotes the zero vector. After recovery, the server performs weighted averaging across the recovered models and gradients for aggregation, and thus obtains the updated global model and gradients.

This recovery mechanism ensures that the model of each device is fully aligned with the structure of the global model, enabling consistent aggregation across heterogeneous devices. {In particular, this recovery step prevents the local lightweighting from permanently excluding any Gaussian point from training: although a device prunes part of its points during local training, the recovery reinstates them from the global model before aggregation. Hence, even if a point is pruned by all devices in the same round, the recovery preserves its parameters intact through the zero-gradient inactive entries, so a temporarily low importance score leads at worst to a missed update in that round rather than an irreversible removal. Whether a point is ultimately densified or pruned is instead governed by the standard server-side adaptive density control, which operates on the aggregated global gradient field rather than on any single device's transient local score.} These points remain eligible for future updates, which avoids the irreversible loss of structural or semantic information in the scene. Moreover, by restoring the full gradient field across all Gaussian points (including zero-filled gradients for inactive points), the recovery mechanism supports the later procedures (i.e., density control), which rely on global gradient information to refine and regenerate scene geometry across rounds.

\section{Experiment Results}
In this section, we present experimental results to evaluate the effectiveness of the proposed framework.

\vspace{-0.1cm}
\subsection{Experimental Setup} \label{Experimental Setup}
\subsubsection{System Configurations} We conduct all experiments in a simulated environment using a single NVIDIA A100 GPU. To emulate the computation characteristics of edge devices, we profile GPU-specific computational parameters through repeated measurements on an NVIDIA RTX 3090 using the NVIDIA Nsight Profiler~\cite{Nsight2023}. These profiled parameters are then used to simulate on-device training behaviors. The proposed design and all benchmark schemes are simulated under this unified simulation setup.

\subsubsection{Learning Datasets} We adopt a real-world large-scale dataset from Mill19 \cite{Turki_2022_CVPR}, which captures two outdoor environments using drone cameras: (1) the \emph{Building} scene: contains 1,940 RGB images at a resolution of $4608 \times 3456 $; and (2) the \emph{Rubble} scene: contains 1,678 RGB images at the same resolution. The dataset also provides accurate per-frame intrinsic and extrinsic camera parameters. {We also adopt (3) the \emph{Polytech} scene from the UrbanScene3D dataset~\cite{UrbanScene3D}, which contains 999 RGB images at a resolution of $5472 \times 3648$.} To simulate a realistic setting, we assign multi-view images to devices in an unbalanced manner, such that each device receives a different number of images with varying scene coverage. Following the strategy used in Fed3D-GS~\cite{fed3Dgs}, we first randomly sample a viewpoint anchor for each device within the scene, and then assign a random number of images captured from nearby camera poses around the anchor. This results in heterogeneous data distributions across devices.

\subsubsection{Implementation}
Experiments are implemented using the PyTorch deep learning framework. Our federated 3D-GS learning framework is developed based on the Flower framework~\cite{beutel2020flower}, while the local training is built based on the official 3D-GS implementation \cite{3DGS}. To ensure fair comparison and reproducibility, all local training hyperparameters are kept consistent with the official 3D-GS. A key modification is that the density control mechanism is disabled during local training and instead is executed centrally on the server during global model aggregation, as presented in Section~\ref{learning_pipeline}. For the federated learning setup, we consider $K=5$ edge devices, each performing $M=150$ local training epochs per communication round, with a total of $N=200$ global communication rounds. 

In the GPU memory model, we assume 32-bit floating-point precision, leading to 4 Bytes per parameter and gradient, i.e., $q_{\text{para}} = q_{\text{opt}} = 4$. The dataset is downsampled by a factor of four, resulting in $|{\boldsymbol{I}}| = 1152 \times 864 = 9.95 \times 10^5$ pixels per image. {The \emph{Polytech} scene is downsampled by the same factor, giving $|{\boldsymbol{I}}| = 1368 \times 912 = 1.25 \times 10^6$ pixels per image.} We adopt a spherical harmonic degree of $v = 3$, consistent with the 3D-GS configuration \cite{3DGS}, yielding 59 parameters per Gaussian point and corresponding optimizer state (i.e., $z_{\text{para}} = z_{\text{opt}} = 59$). We also set $z_{\text{temp}} = 7$ for intermediate tensors and $z_{\boldsymbol{\psi}} = 51$ for camera parameters. For memory-constrained edge devices, we set the maximum GPU memory to $6~\text{Gigabytes}$, with a base memory consumption of $1.5~\text{Gigabytes}$. According to official NVIDIA documentation \cite{nvidia_ampere_tuning_guide,nvidia_ampere_ga102_v2,nvidia_ampere_ga106}, the GPU supports up to $3.558\times 10^{13}~\text{FLOPs/s}$ computation throughput, a peak memory bandwidth of $9.36 \times 10^{11}~\text{Bytes/s}$, and 
125,952 threads for parallel execution. The computation efficiency coefficients are set as $\epsilon=0.38$ and $\tilde{\epsilon}=0.74$. In the computation latency model, we set $N_{\text{FLOPs}}= 657408~\text{FLOPs}$ and $B= 550400~\text{Bytes}$ for the splatting stage, and $\tilde{N}_{\text{FLOPs}}=12~\text{MFLOPs}$, $\tilde{B}=0.35~\text{MBytes}$ for the rendering stage. The scale factor parameters are set as $\phi_1 = 1.67$ and $\phi_2 = 0.7$. 

In the communication model, we utilize real-world 5G measurement datasets~\cite{uplink5G,downlink5G} to simulate practical large-scale wireless conditions. Specifically, the downlink communication rate is obtained from a commercial 5G dataset reported in~\cite{downlink5G}, while uplink communication rates are derived from two distinct scenarios in the 5G measurement dataset~\cite{uplink5G}, corresponding to mobile (driving) and static user environments. During simulation, each device independently samples one of the two uplink environments as its communication setting.

\subsubsection{Benchmark Schemes} We consider six benchmark schemes for performance comparison.
\begin{itemize}
    \item \textbf{3D-GS \cite{3DGS}:} This scheme represents the centralized training setup, where all training images and their corresponding camera parameters are aggregated at a single computing node. The standard 3D-GS training pipeline is executed without any data or model distribution.
    \item \textbf{Fed3D-GS \cite{fed3Dgs}:} This state-of-the-art benchmark performs independent 3D-GS training on each device, where the entire local model is trained without any intermediate model exchange. After local training is completed, each device uploads its model to the server in a one-shot communication manner. The server then conducts a refinement stage to construct the global model, which also follows the standard 3D-GS training pipeline (i.e., splatting and rendering). As such, the communication latency consists of only a single communication round. However, the overall computation latency includes both local training and server-side retraining, which can be also estimated using our latency model.
    \item \textbf{Full memory load with recovery design:} Each device selects the maximum allowable number of Gaussian points determined by its GPU memory and latency constraints in each communication round, i.e., $U_k^{(n)}=\hat{U}_k,~\forall n\in\mathcal{N}$, and the recovery mechanism is used to restore the model and gradients on the server.
    \item \textbf{Lightweighting only design:} Each device first selects the optimal number of Gaussian points $U_k^{(n)*}$ by solving problem \eqref{maxratio}, and then performs local training. The resulting local models are uploaded to the server and directly merged without parameter averaging, i.e., $\mathcal{G}^{(n+1)} = \bigcup_{i=1}^{K} \mathcal{G}_i^{(n)}$. 
    \item \textbf{Full model training design:} We deactivate the memory and latency constraints of edge device in this scheme. At each communication round, each device locally trains on the complete set of Gaussian points based on its dataset and uploads the entire model to the server, where parameter averaging is applied to update the global model.
    \item {\textbf{DOGS \cite{DOGS}:} This distributed 3D-GS benchmark recursively partitions a large-scale scene into balanced blocks and trains the corresponding local models in parallel through Gaussian consensus. We retain its original training and consensus framework without algorithmic modifications. To align its local training budget with our setting, we set the number of local training epochs between two consecutive consensus operations to $M=150$. We further add a latency-accounting module based on the computation and transmission latency models in Section~\ref{model_lightweighting} to evaluate its end-to-end training latency for comparison.}

\end{itemize}

\subsubsection{Performance Metrics} We adopt the following three widely adopted performance metrics in 3D reconstruction tasks: a) peak signal-to-noise ratio (PSNR),  which measures pixel-wise reconstruction error, where higher values indicate better reconstruction quality; b) SSIM \cite{SSIM}, which evaluates the similarity between ground truth and rendered images, where a higher SSIM indicates better preservation of structural content; and c) learned perceptual image patch similarity (LPIPS) \cite{LPIPS}, which assesses perceptual similarity, with lower values indicating closer to the ground truth.

\vspace{-0.25cm}

\subsection{Simulation Results} \label{Simulation Results}
\subsubsection{Convergence Behavior} 
\begin{figure*}[!t]
    \centering
    
    \subfloat[Building scene\label{fig:loss_building}]{
        \includegraphics[width=0.275\textwidth]{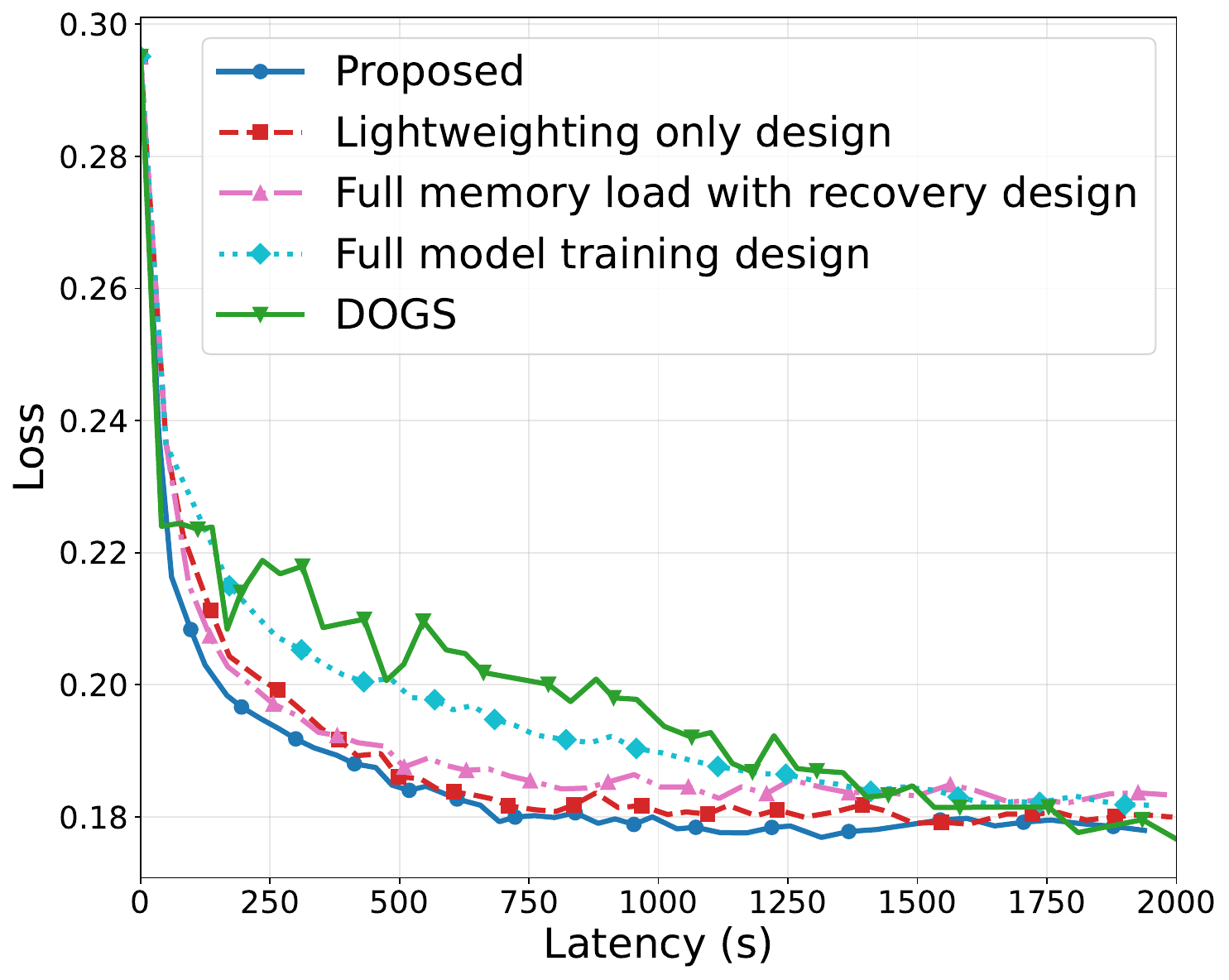}
    } \hfill
    \subfloat[Rubble scene\label{fig:loss_rubble}]{
        \includegraphics[width=0.275\textwidth]{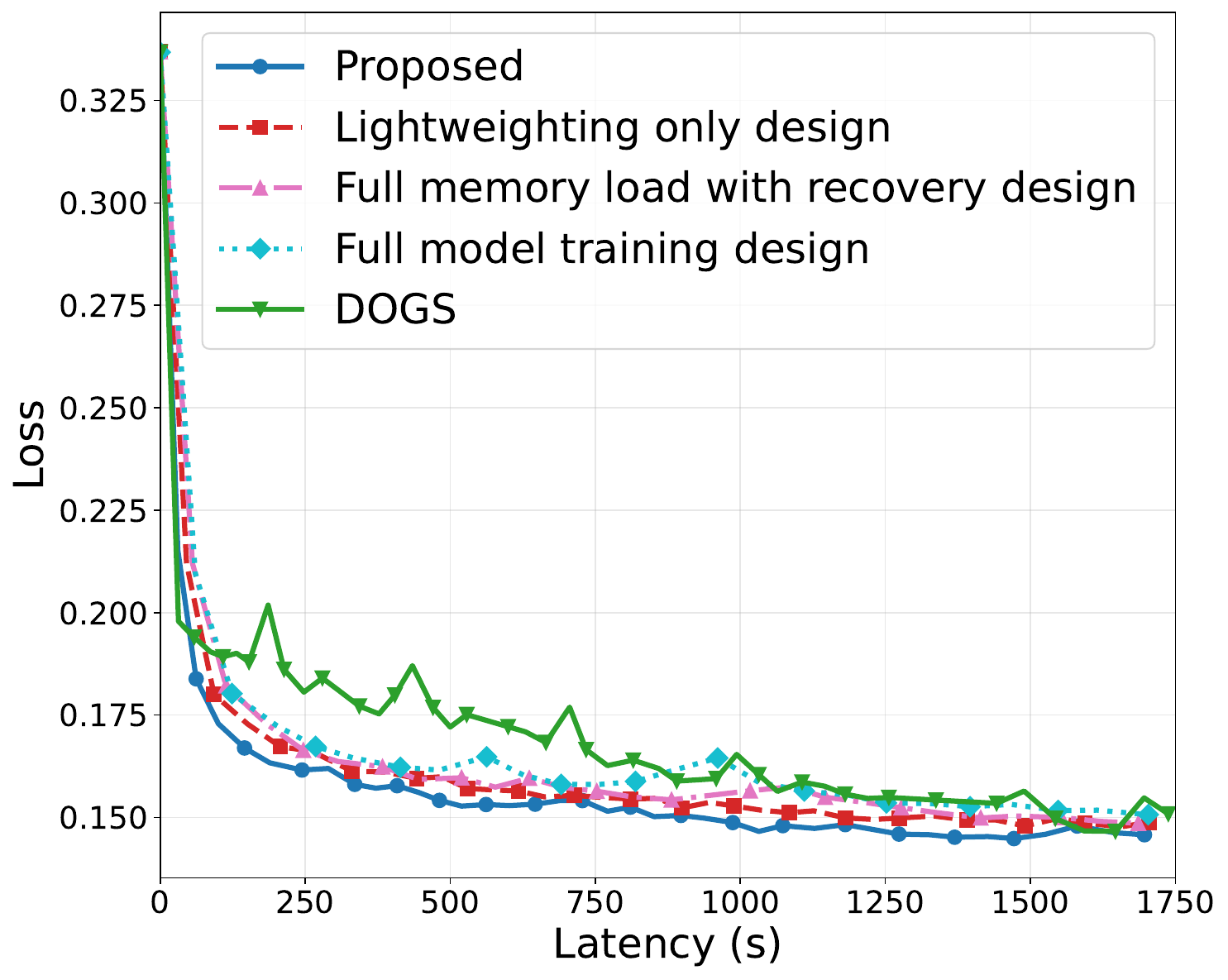}
    } \hfill
    \subfloat[Polytech scene\label{fig:loss_polytech}]{
        \includegraphics[width=0.275\textwidth]{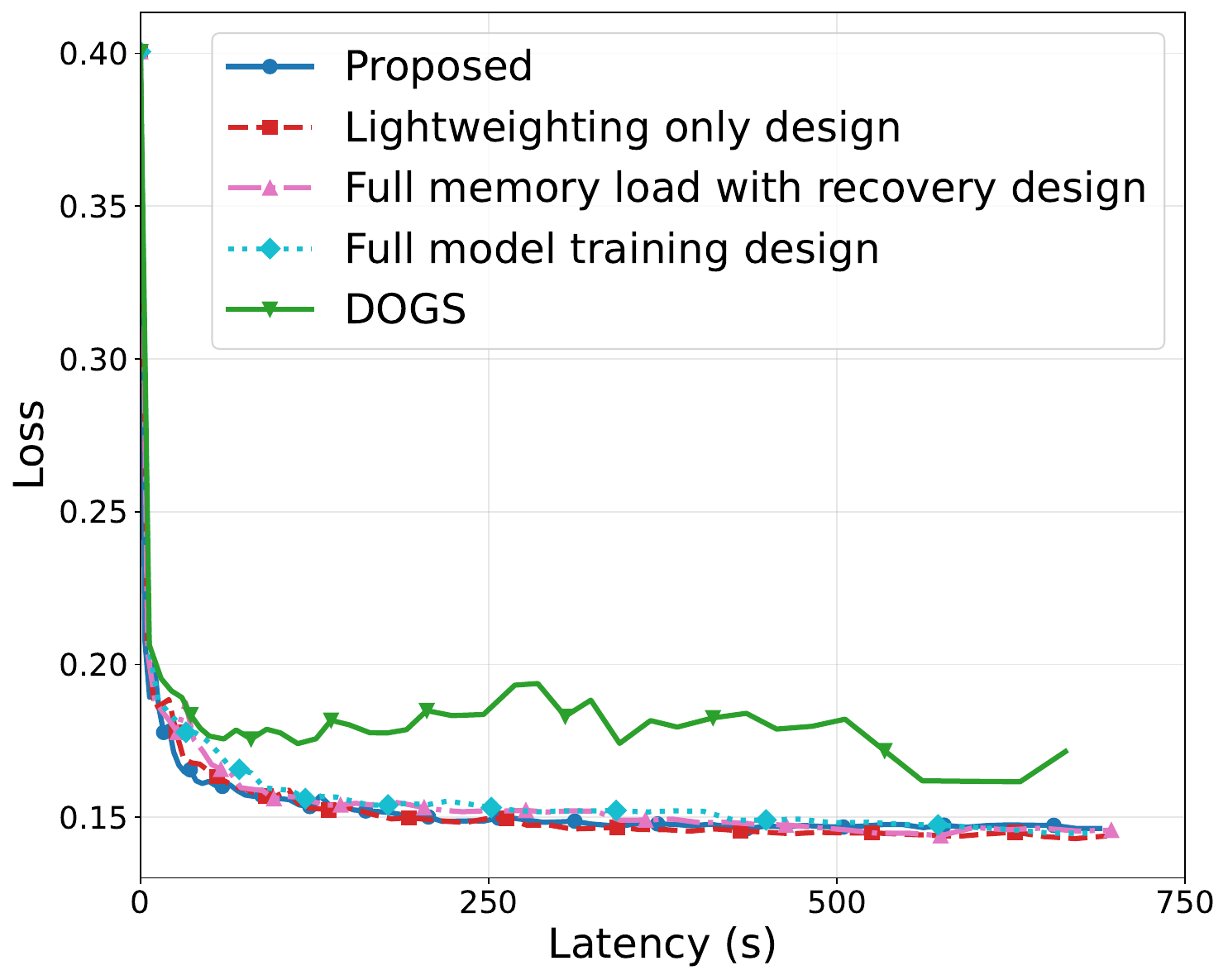}
    }
    \caption{Comparison of training loss convergence versus latency on (a) the \emph{Building} scene, (b) the \emph{Rubble} scene, and (c) the \emph{Polytech} scene.}
    \label{convergence}
\vspace{-1.5em}
\end{figure*}

Fig.~\ref{convergence} illustrates the convergence behavior in terms of the training loss versus the overall latency in \emph{Building} and \emph{Rubble} scenes. The proposed method consistently demonstrates the fastest convergence, achieving rapid loss reduction under constrained latency budgets. Specifically, the full model training design shows the slowest convergence due to significant per-round training and uplink transmission overhead, as each device must train and upload a complete 3D-GS model. While this avoids performance loss from pruning and yields smooth convergence, its excessive latency makes it impractical for deployment at wireless edge. The full memory load training with recovery design leverages the maximum allowed memory capacity for local training, which leads to faster convergence than the full model scheme. However, it still suffers from notable overhead in local training time compared to the proposed method. In comparison, the lightweighting only design adopts the same lightweighting strategy as our method and exhibits similar early-stage convergence. Nevertheless, due to the absence of a recovery mechanism, it suffers from performance degradation during model aggregation. As training progresses, our method progressively outperforms the lightweighting only design. Moreover, since the lightweighting only design directly merges local models without parameter averaging, the global model size grows rapidly, increasing downlink communication latency. {Furthermore, compared with the DOGS scheme, the proposed method also achieves faster convergence under the same latency budget. This is because DOGS trains full block-wise models and periodically enforces Gaussian consensus across blocks without resource-aware lightweighting, which incurs heavier per-round local training and synchronization overhead and thus slows down its training.} These results validate that our design achieves faster convergence with significantly reduced training latency cost.

{For the \emph{Polytech} scene, the loss curves exhibit a similar convergence trend. The proposed method rapidly reduces the training loss under constrained latency budgets and reaches a stable loss level comparable to the lightweighting only, full memory load with recovery, and full model training designs, whereas DOGS exhibits a higher and more fluctuating loss over most of the latency range.}

\subsubsection{Learning Performance}
\begin{figure*}[!t]
    \centering
    
    \subfloat[PSNR versus total latency\label{fig:subfig_a_building}]{
        \includegraphics[width=0.275\textwidth]{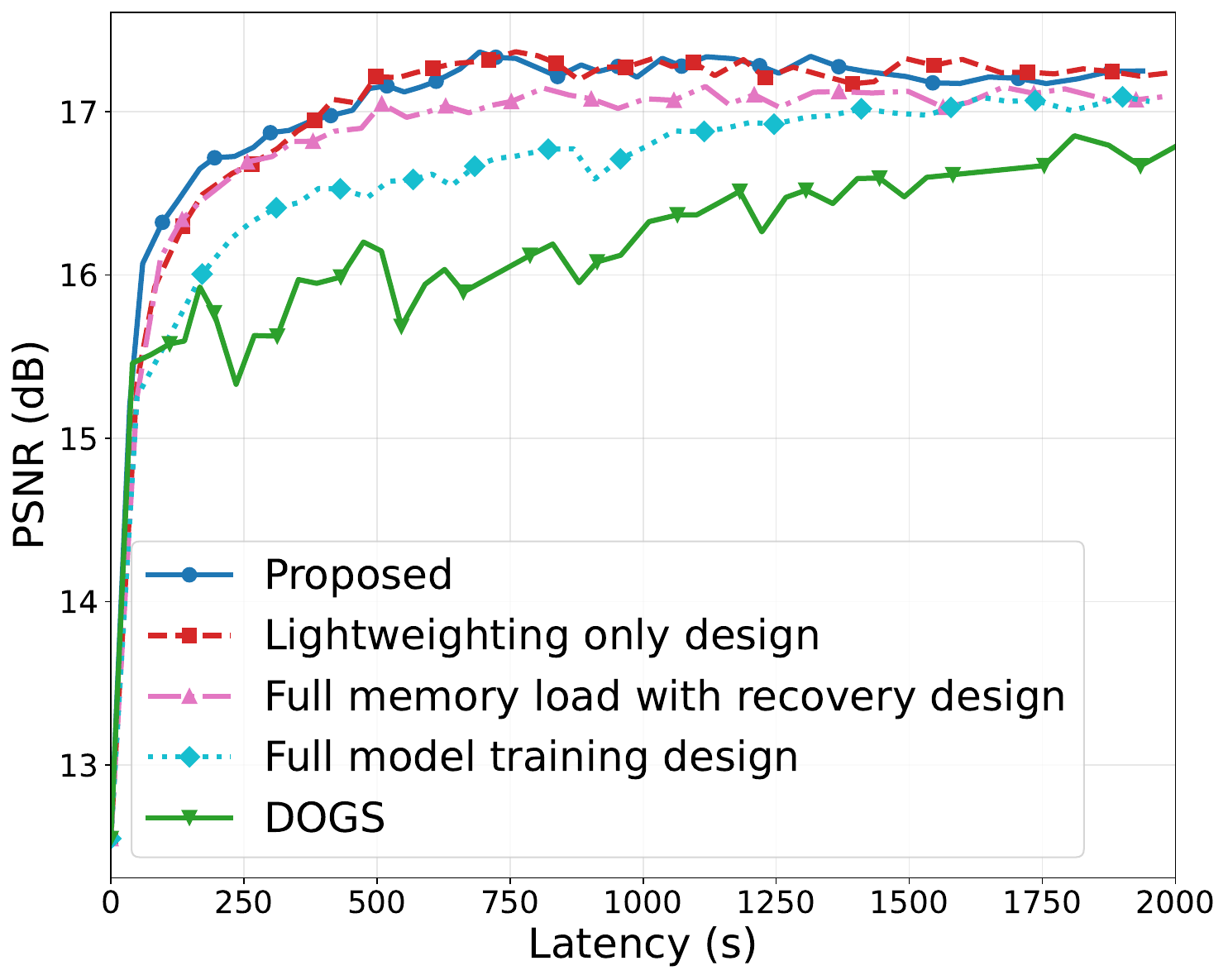}
    } \hfill
    \subfloat[SSIM versus total latency\label{fig:subfig_b_building}]{
        \includegraphics[width=0.275\textwidth]{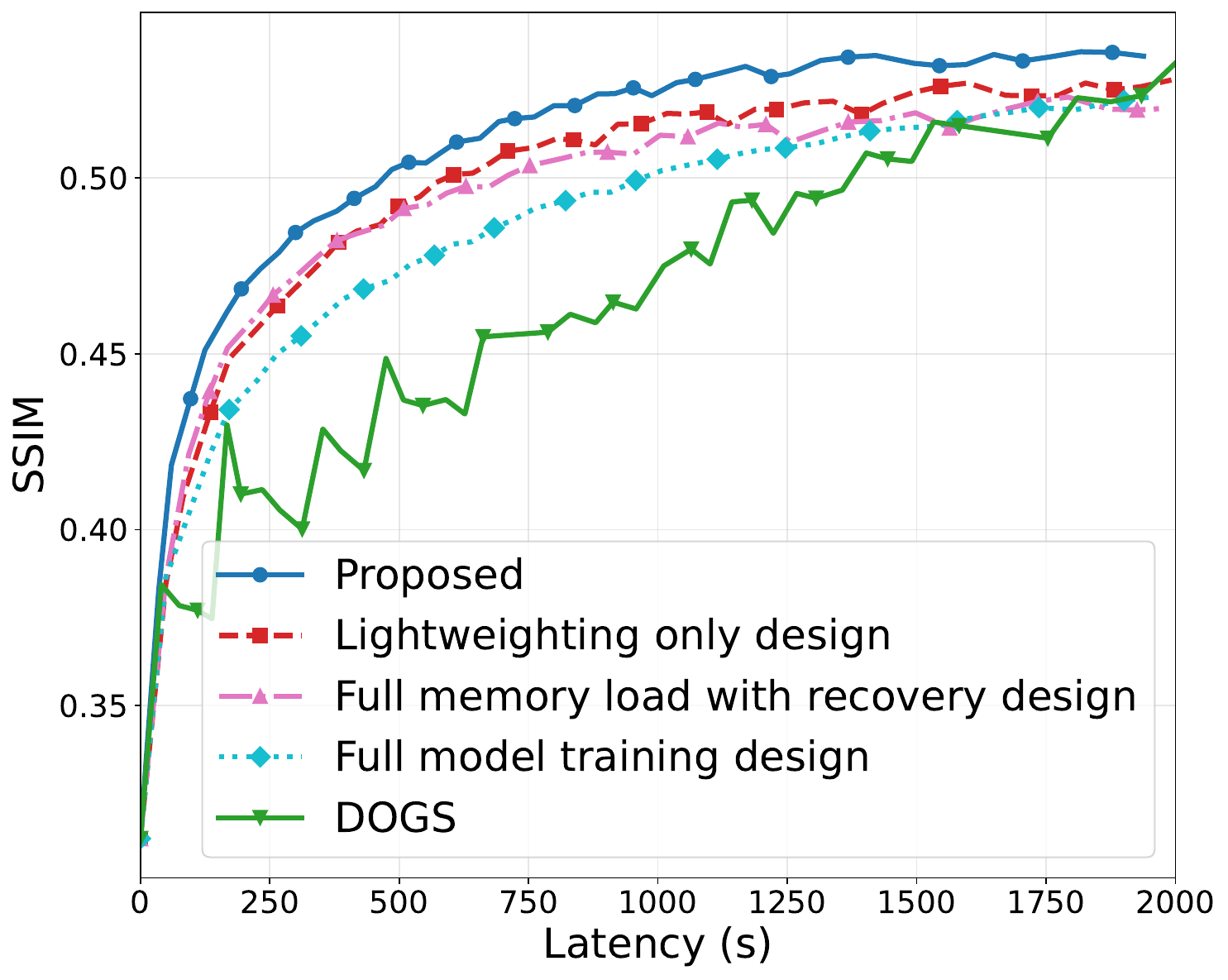}
    } \hfill
    \subfloat[LPIPS versus total latency\label{fig:subfig_c_building}]{
        \includegraphics[width=0.275\textwidth]{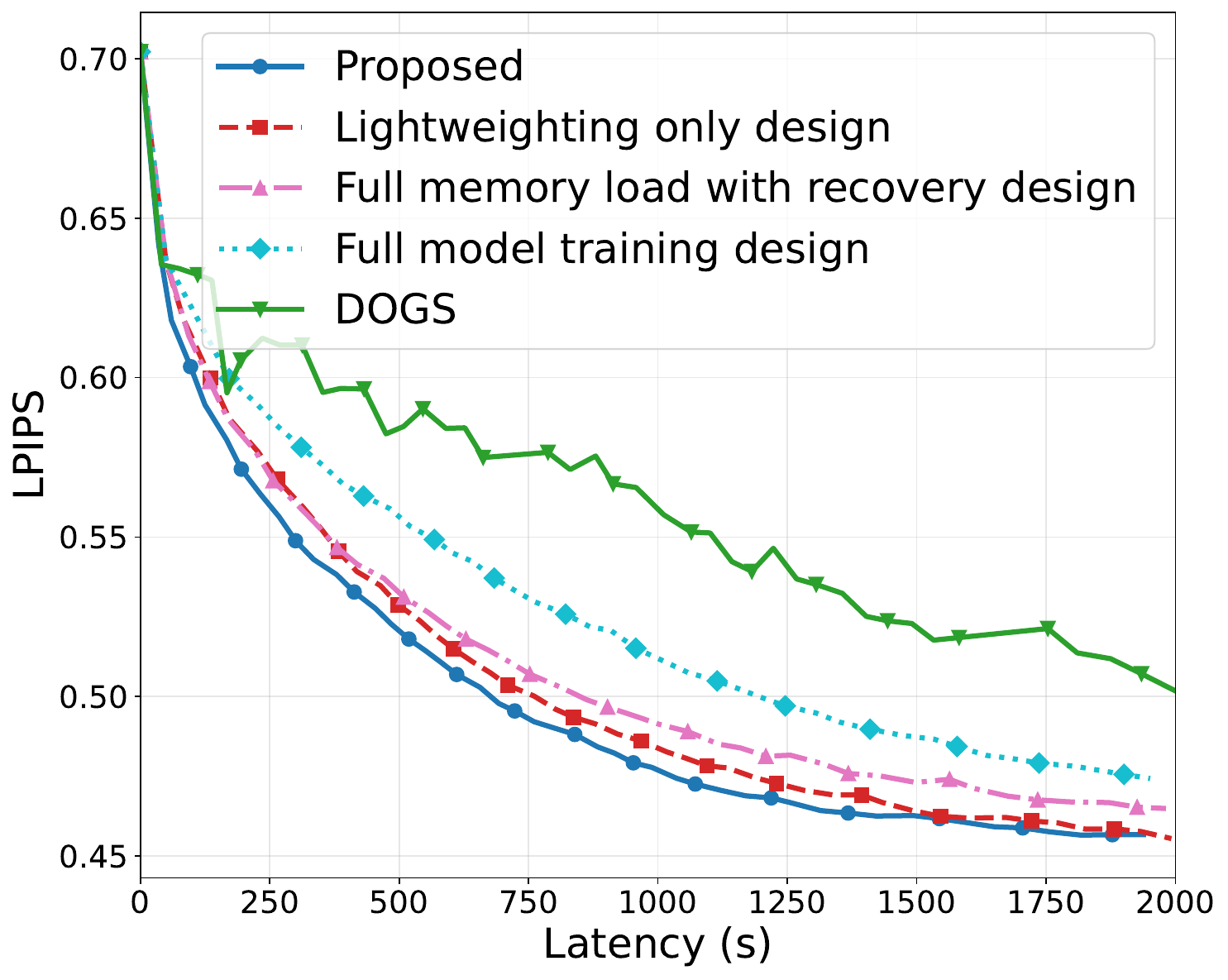}
    }
    \caption{Comparison of different training designs in terms of PSNR, SSIM, and LPIPS w.r.t. total latency in the \emph{Building} scene.}
    \label{test_building}
\vspace{-1.5em}
\end{figure*}
\begin{figure*}[!t]
    \centering
    
    \subfloat[PSNR versus total latency\label{fig:subfig_a_rubble}]{
        \includegraphics[width=0.275\textwidth]{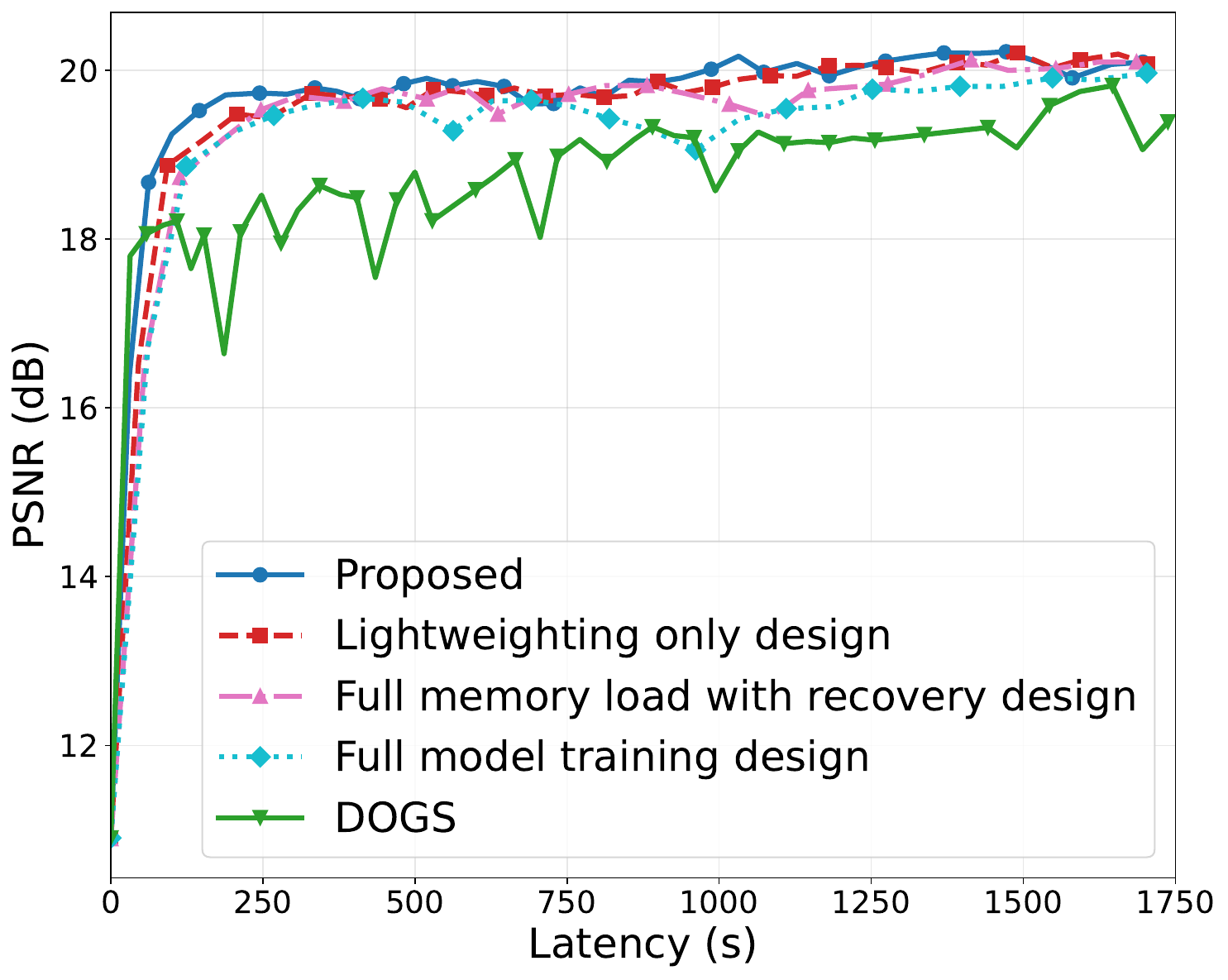}
    } \hfill
    \subfloat[SSIM versus total latency\label{fig:subfig_b_rubble}]{
        \includegraphics[width=0.275\textwidth]{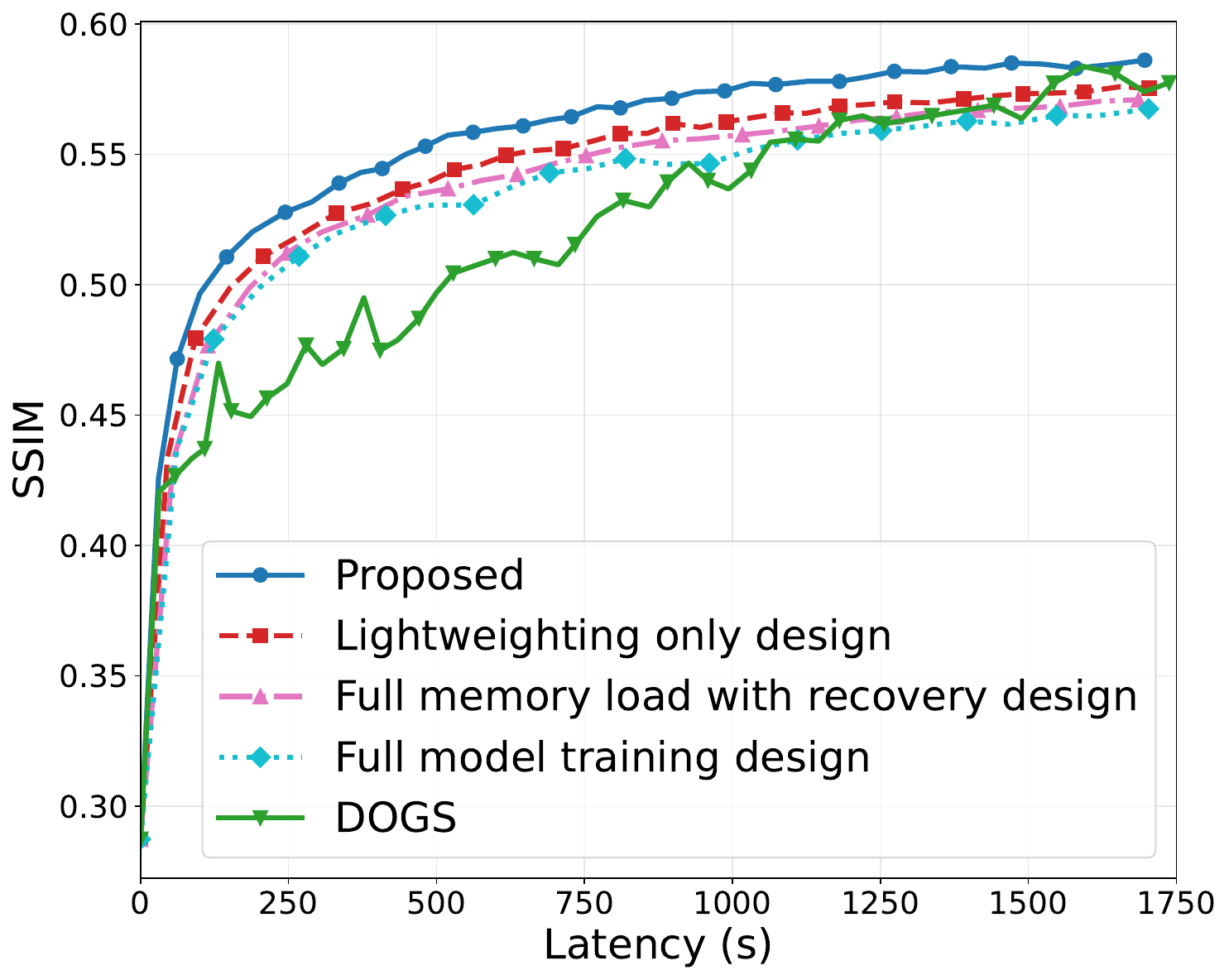}
    } \hfill
    \subfloat[LPIPS versus total latency\label{fig:subfig_c_rubble}]{
        \includegraphics[width=0.275\textwidth]{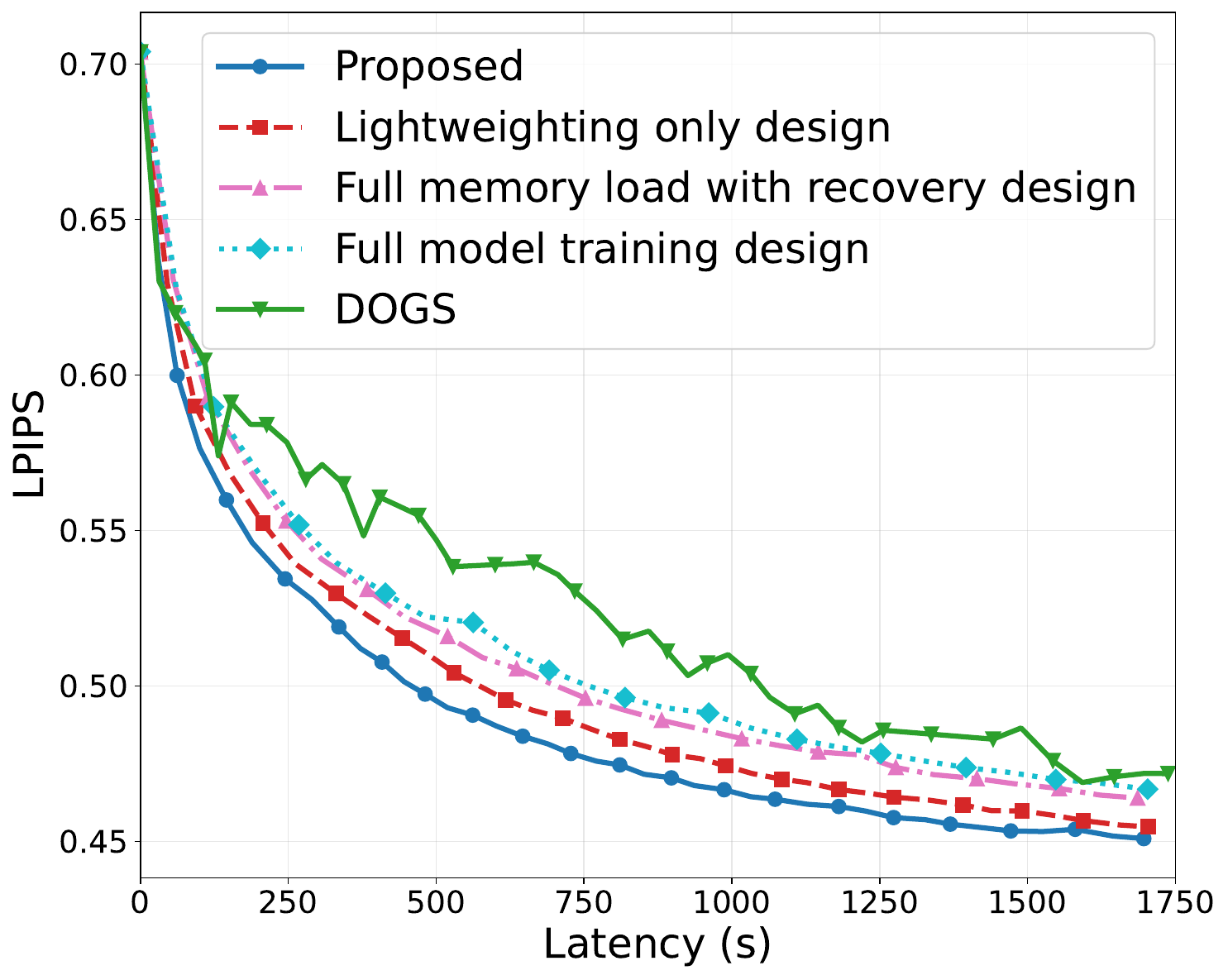}
    }
    \caption{Comparison of different training designs in terms of PSNR, SSIM, and LPIPS w.r.t. total latency in the \emph{Rubble} scene.}
    \label{test_rubble}
\vspace{-1.2em}
\end{figure*}
\begin{figure*}[!t]
    \centering
    
    \subfloat[PSNR versus total latency\label{fig:subfig_a_polytech}]{
        \includegraphics[width=0.275\textwidth]{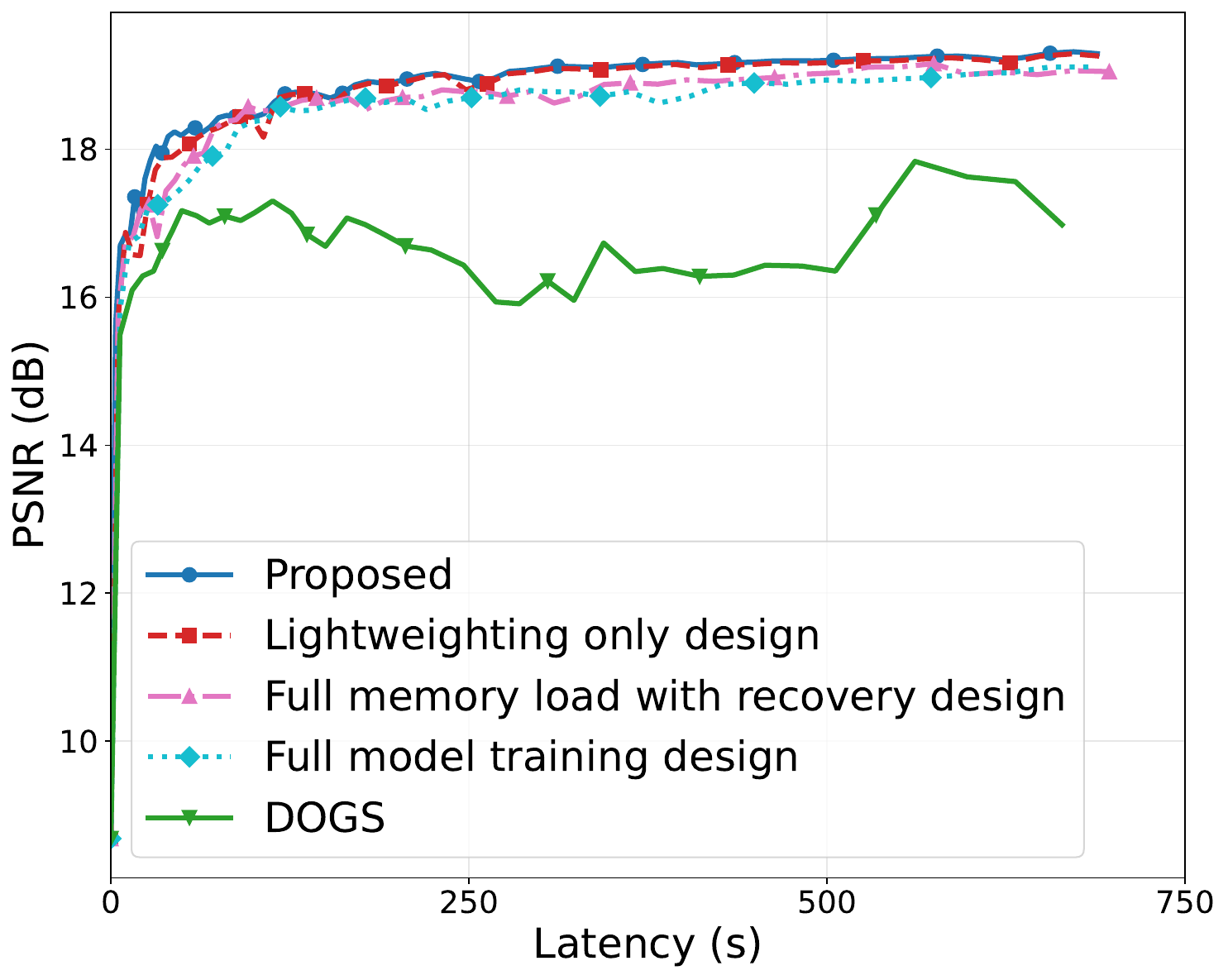}
    } \hfill
    \subfloat[SSIM versus total latency\label{fig:subfig_b_polytech}]{
        \includegraphics[width=0.275\textwidth]{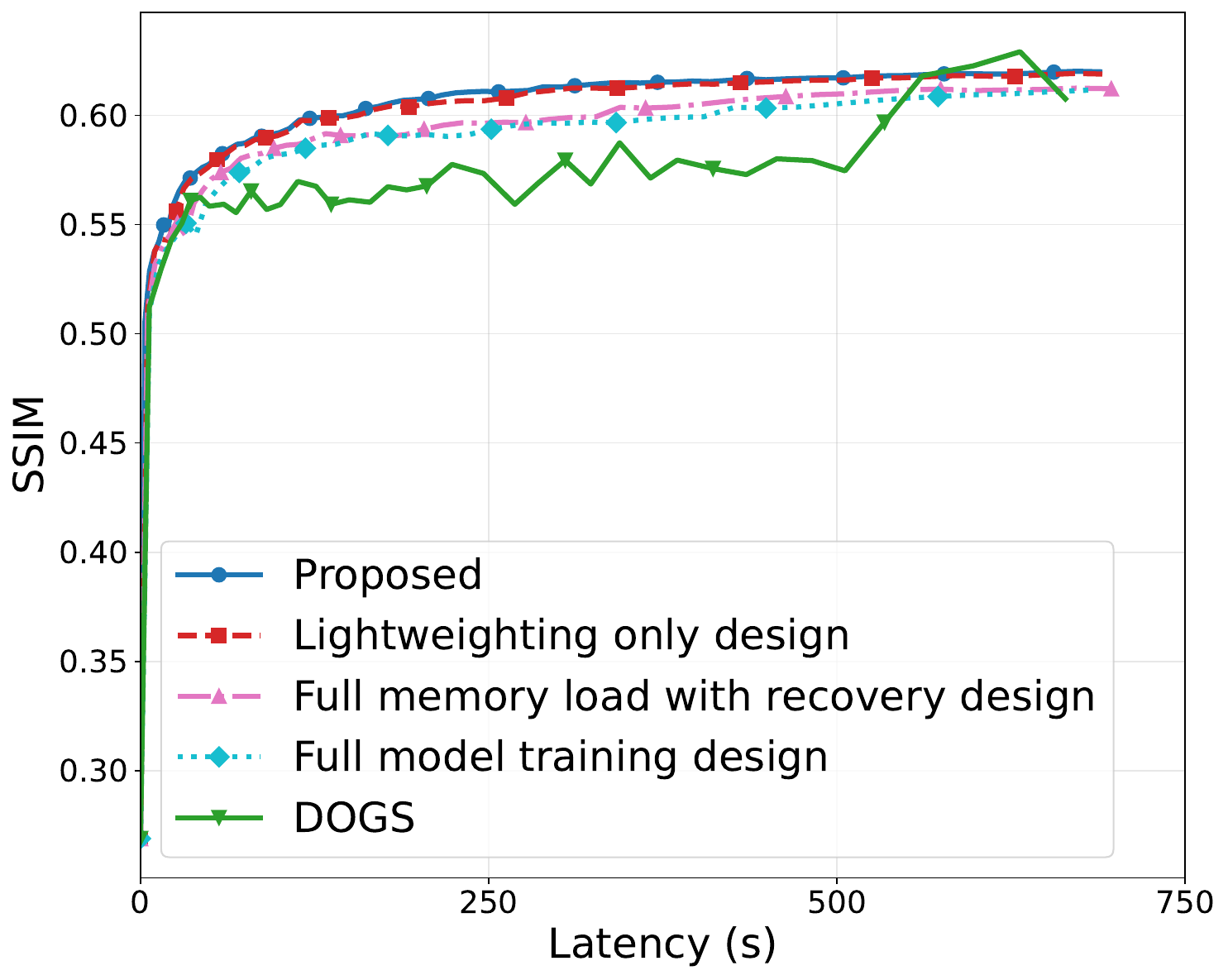}
    } \hfill
    \subfloat[LPIPS versus total latency\label{fig:subfig_c_polytech}]{
        \includegraphics[width=0.275\textwidth]{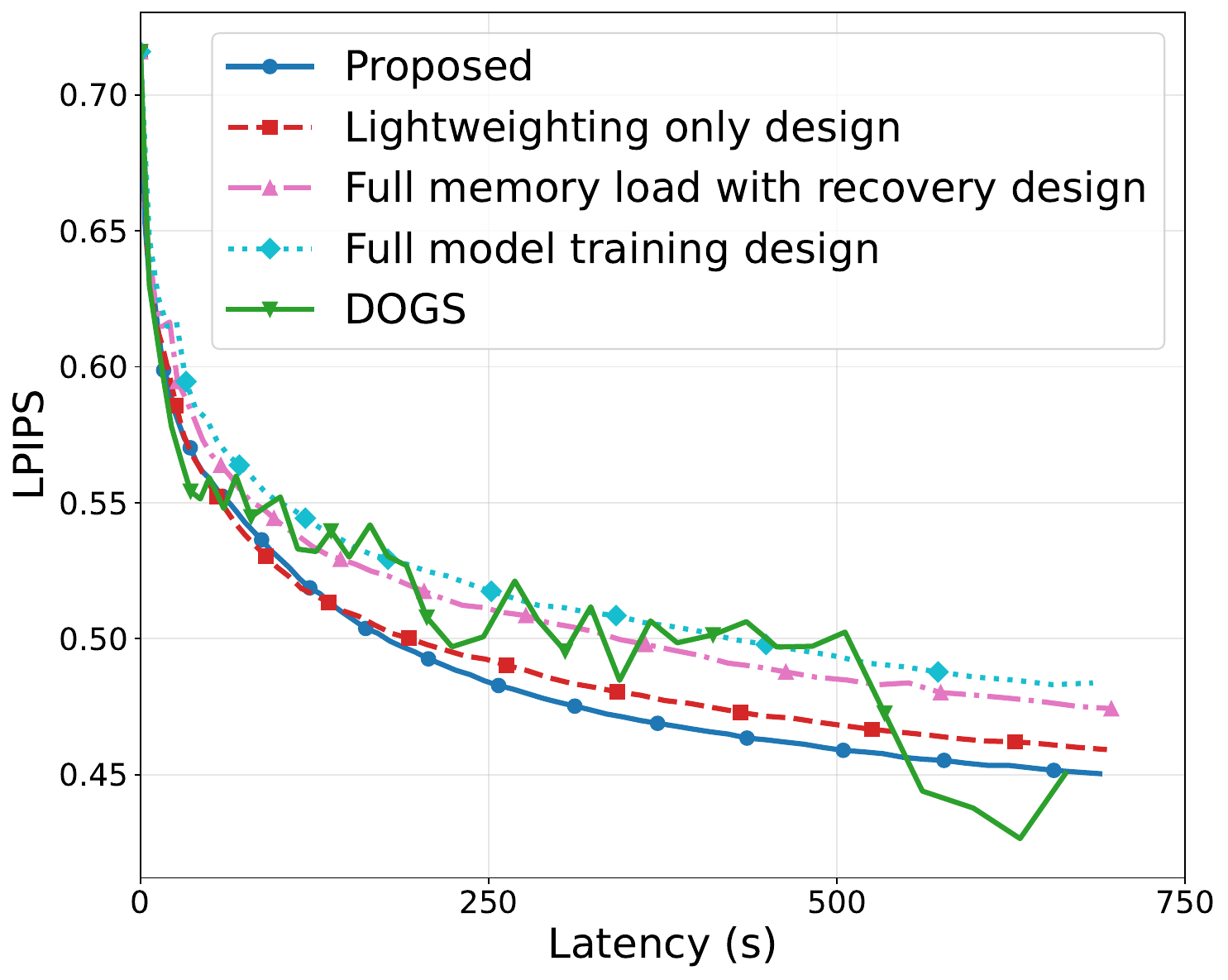}
    }
    \caption{Comparison of different training designs in terms of PSNR, SSIM, and LPIPS w.r.t. total latency in the \emph{Polytech} scene.}
    \label{test_polytech}
\vspace{-1.2em}
\end{figure*}
Fig.~\ref{test_building} and Fig.~\ref{test_rubble} present the test performance in terms of PSNR, SSIM, and LPIPS versus total latency. It is observed that the proposed method outperforms all baseline schemes across all three metrics, achieving higher PSNR and SSIM, and lower LPIPS throughout the training process. In the early training stages, our method and the lightweighting only design exhibit comparable performance due to the same model lightweighting strategy. However, the lightweighting only design exhibits slower improvement in reconstruction quality during training compared to the proposed method, particularly in terms of perceptual similarity as reflected by LPIPS. In contrast, the proposed method achieves faster and more consistent improvement in all metrics. Compared with the full model training and full memory load with recovery designs, in which those Gaussian points that contribute little to the rendering also spend substantial computation and communication resources, our approach better balances training cost and visual fidelity by selecting the most important points under constrained GPU memory and latency. {Compared with the DOGS scheme, the proposed method achieves higher PSNR and SSIM and lower LPIPS under the same total latency in both scenes.}

{It is observed from Fig.~\ref{test_polytech} that the proposed method achieves the best overall reconstruction quality on the \emph{Polytech} scene. Under a small latency budget, it rapidly improves all three quality metrics, achieving the highest PSNR and SSIM and the lowest LPIPS over most of the evaluated latency range, consistent with the trend observed on the \emph{Building} and \emph{Rubble} scenes. In contrast, DOGS shows lower PSNR and SSIM for most of the range, with substantially larger fluctuations.}

\begin{table}[t]
\centering
\caption{Comparison of Reconstruction Quality With Fed3D-GS.}
\label{comparison_fed3dgs}
\setlength{\tabcolsep}{3pt} %
\renewcommand{\arraystretch}{1.2} %
\begin{tabular}{l|ccc|ccc}
\toprule
\multirow{2}{*}{Scenes} & \multicolumn{3}{c|}{\textbf{Building}} & \multicolumn{3}{c}{\textbf{Rubble}} \\
 & PSNR $\uparrow$ & SSIM $\uparrow$ & LPIPS $\downarrow$  & PSNR $\uparrow$ & SSIM $\uparrow$ & LPIPS $\downarrow$ \\
\midrule
Fed3D-GS          & 17.21 & 0.567 & 0.375 & 19.18 & 0.552 & 0.429 \\
Proposed              & \textbf{17.91} & 0.554 & \textbf{0.363} & \textbf{20.74} & \textbf{0.628} & 0.438 \\
\bottomrule
\end{tabular}
\vspace{-1em}
\end{table}

\begin{table}[t]
  \centering
  \setlength{\tabcolsep}{2pt}    
  \renewcommand{\arraystretch}{1.1} 
  \caption{Comparison of Total Latency (in Hours),  Number of Gaussian Points in local model, and Peak GPU Memory Consumption (in GB) for One Device after 30,000 Training Epochs.}
  \label{comparison_efficiency}
  \begin{tabular*}{\columnwidth}{@{\extracolsep{\fill}} l | c c c | c c c @{}}
    \toprule
    \multirow{2}{*}{Method}
      & \multicolumn{3}{c|}{\textbf{Building}}
      & \multicolumn{3}{c}{\textbf{Rubble}} \\
      & Latency & Points & Memory
      & Latency & Points & Memory \\
    \midrule
    3D-GS    & 9.27  & 4330717   & 7.18  & 7.71  & 3154812   & 6.59  \\
    Fed3D-GS & 1.83  & 1967079   & 5.93   & 1.69  & 1001575   & 4.68  \\
    Proposed & \textbf{1.51} & \textbf{1358773} & \textbf{5.69}
             & \textbf{1.39} & 1260379 & 4.81 \\
    \bottomrule
  \end{tabular*}
  \vspace{-1.5em}
\end{table}

Table~\ref{comparison_fed3dgs} compares the proposed method with the state-of-the-art federated 3D-GS learning framework Fed3D-GS \cite{fed3Dgs} on the \emph{Building} and \emph{Rubble} scenes. It is observed that our method achieves better performance in PSNR and LPIPS in the \emph{Building} scene, while achieves a significant PSNR improvement and higher SSIM in the \emph{Rubble} scene. Table~\ref{comparison_efficiency} compares the latency and GPU memory consumption of different training designs for one device after 30,000 training epochs. It is observed that the proposed design achieves the lowest training latency, Gaussian point count, and the peak memory consumption in the \emph{Building} scene, while also maintains the lowest training latency in the \emph{Rubble} scene, with slightly higher model size and memory usage compared to Fed3D-GS. These results highlight the advantages of our method in improving training efficiency while maintaining high reconstruction quality under resource constraints.

\subsubsection{Robustness to Heterogeneous Devices}

{
\begin{figure}[t]
    \centering
    \subfloat[PSNR\label{fig:subfig_a_camhetero}]{
        \includegraphics[width=0.40\columnwidth]{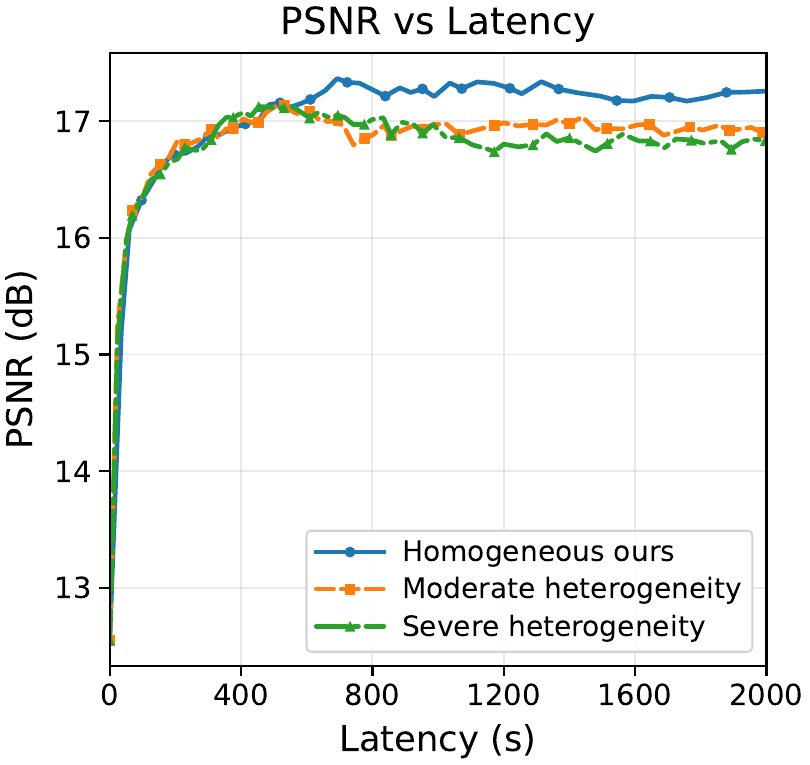}
    } \hfill
    \subfloat[SSIM\label{fig:subfig_b_camhetero}]{
        \includegraphics[width=0.40\columnwidth]{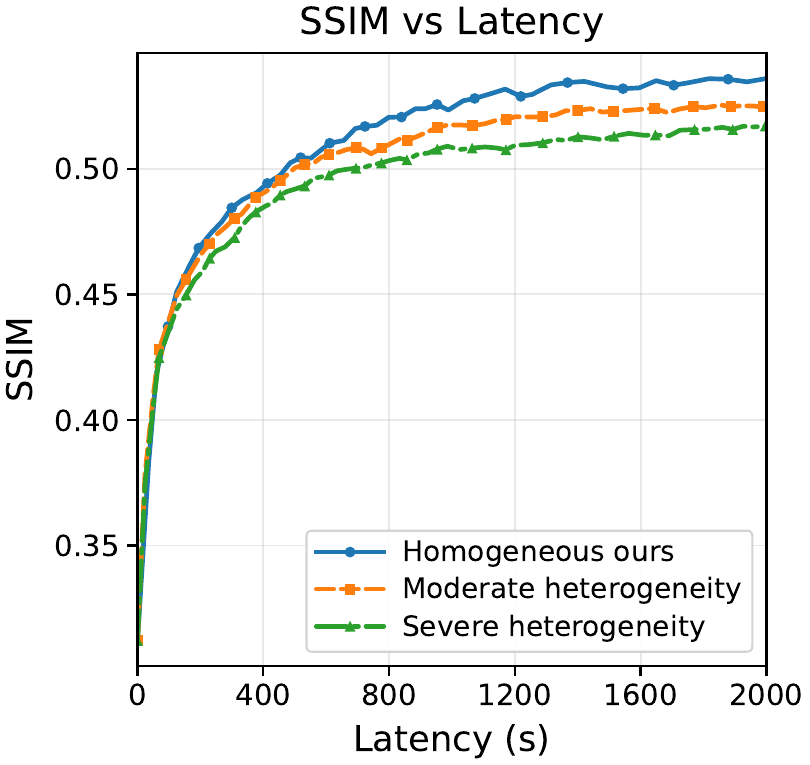}
    } \\[0.1em]
    \subfloat[LPIPS\label{fig:subfig_c_camhetero}]{
        \includegraphics[width=0.40\columnwidth]{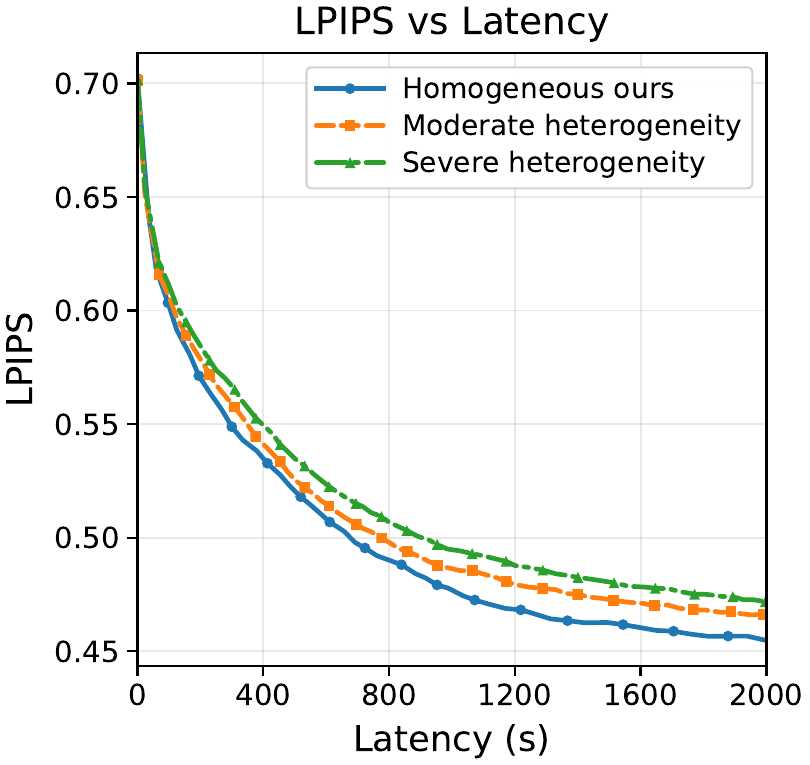}
    } \hfill
    \subfloat[Training loss\label{fig:subfig_d_camhetero}]{
        \includegraphics[width=0.40\columnwidth]{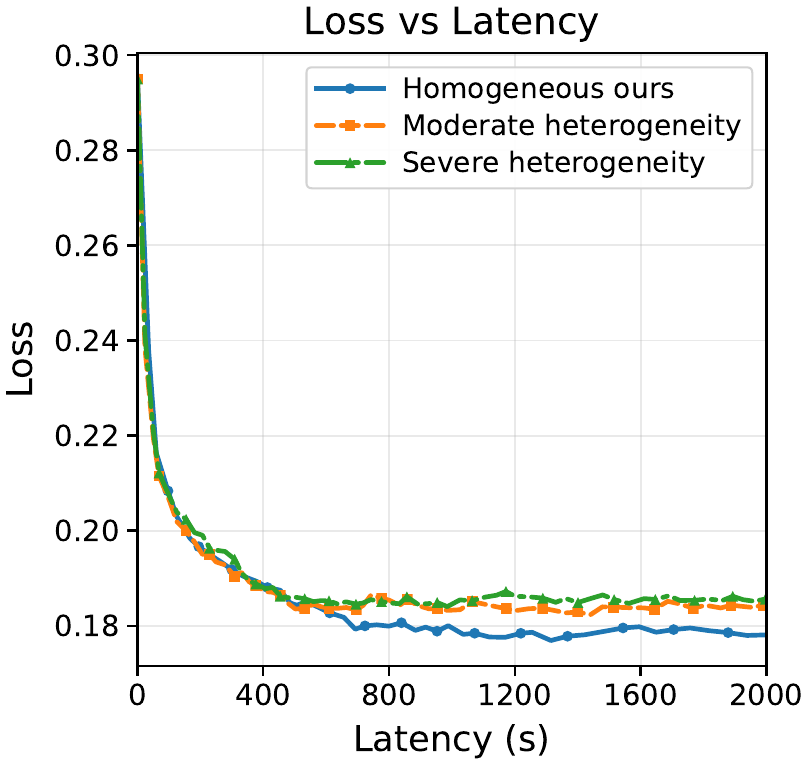}
    }
    \caption{Rendering quality and training loss w.r.t. total latency on the \emph{Building} scene under moderate/severe camera heterogeneity.}
    \label{test_camera_heterogeneity}
    \vspace{-1.2em}
\end{figure}

\begin{table}[t]
  \centering
  \setlength{\tabcolsep}{4pt}
  \renewcommand{\arraystretch}{1.2}
  \caption{Reconstruction Quality Under Device Heterogeneity on the \emph{Building} Scene.}
  \label{camera_heterogeneity_table}
  {
  \begin{tabular}{l|ccc}
    \toprule
    Setting & PSNR $\uparrow$ & SSIM $\uparrow$ & LPIPS $\downarrow$ \\
    \midrule
    Homogeneous (ours)      & \textbf{17.91} & \textbf{0.554} & \textbf{0.363} \\
    Moderate heterogeneity  & 16.92 & 0.525 & 0.466 \\
    Severe heterogeneity    & 16.88 & 0.518 & 0.471 \\
    \bottomrule
  \end{tabular}}
  \vspace{-1.5em}
\end{table}

To evaluate the robustness of the proposed training framework under heterogeneous acquisition devices, we simulate five federated devices that capture the \emph{Building} scene using cameras of different types and intrinsic parameters. The cameras are heterogeneous along three practical axes, i.e., image resolution, focal length (and thus field of view), and sensor noise, so that the five devices range from a high-quality reference camera to low-resolution, telephoto, and noisy ones. Each device keeps its own camera parameters $\boldsymbol{\psi}_{k,i}$ while jointly reconstructing the same scene. To study different degrees of heterogeneity, we consider two controlled levels, i.e., a moderate level and a more challenging severe level that further enlarges these differences.

We compare the homogeneous-camera setting with the moderate and severe heterogeneity settings in terms of both the training behavior and the converged rendering quality after 30,000 training iterations. It is observed from Fig.~\ref{test_camera_heterogeneity} that the proposed framework remains effective under heterogeneous devices: even when the five devices use inconsistent camera types and intrinsic parameters, the training still converges normally, closely following the homogeneous convergence trajectory in the early stage and reaching a comparable stable plateau. This confirms that mixing heterogeneous cameras does not destabilize local training, recovery, or aggregation, and a high-quality global model can still be learned. It is further observed from Table~\ref{camera_heterogeneity_table} that the framework is robust to the degree of heterogeneity: compared with the homogeneous case, all three metrics degrade only slightly and non-catastrophically, with the PSNR dropping by about $1$~dB even under severe heterogeneity. Moreover, when the heterogeneity is further intensified from the moderate to the severe level, the PSNR decreases slightly, which reflects the stability of the proposed framework as the degree of heterogeneity grows.
}

\subsubsection{Effect of Low-Precision Communication}

\begin{table}[t]
  \centering
  \setlength{\tabcolsep}{4pt}
  \renewcommand{\arraystretch}{1.2}
  \caption{Round-Wise Communication-Precision Ablation on the \emph{Building} Scene.}
  \label{precision_ablation_table}
  {
  \begin{tabular}{l|ccc}
    \toprule
    Communication format & PSNR $\uparrow$ & SSIM $\uparrow$ & LPIPS $\downarrow$ \\
    \midrule
    FP32 (ours) & \textbf{17.91} & \textbf{0.554} & \textbf{0.363} \\
    FP16        & 17.19 & 0.536 & 0.448 \\
    INT8        & 11.03 & 0.290 & 0.766 \\
    \bottomrule
  \end{tabular}}
  \vspace{-1.5em}
\end{table}

{
We compare how transmitting the exchanged 3D-GS parameters at different numerical precisions affects the reconstruction quality. All local training, aggregation, and rendering are kept in FP32, while at each communication round the outgoing Gaussian parameters are converted to FP16, or quantized to 8-bit integers (INT8) through a simple linear mapping, before transmission and converted back to FP32 after reception, so that the ablation reflects the cumulative effect of low-precision communication over the whole training process. It is observed from Table~\ref{precision_ablation_table} that FP16 communication incurs only a modest quality loss, with the PSNR and SSIM decreasing by merely $0.72$~dB and $0.018$ and the LPIPS increasing by only $0.085$ relative to FP32, indicating that FP16 is a practical format for model exchange. In contrast, directly compressing the exchanged models to INT8 degrades the quality substantially, since its much larger rounding error is repeatedly injected into the global and local models and accumulates over training.
}

\subsubsection{Analysis of Latency Overhead}

\begin{figure}[htbp!]
  \centering
  \includegraphics[width=0.85\linewidth]{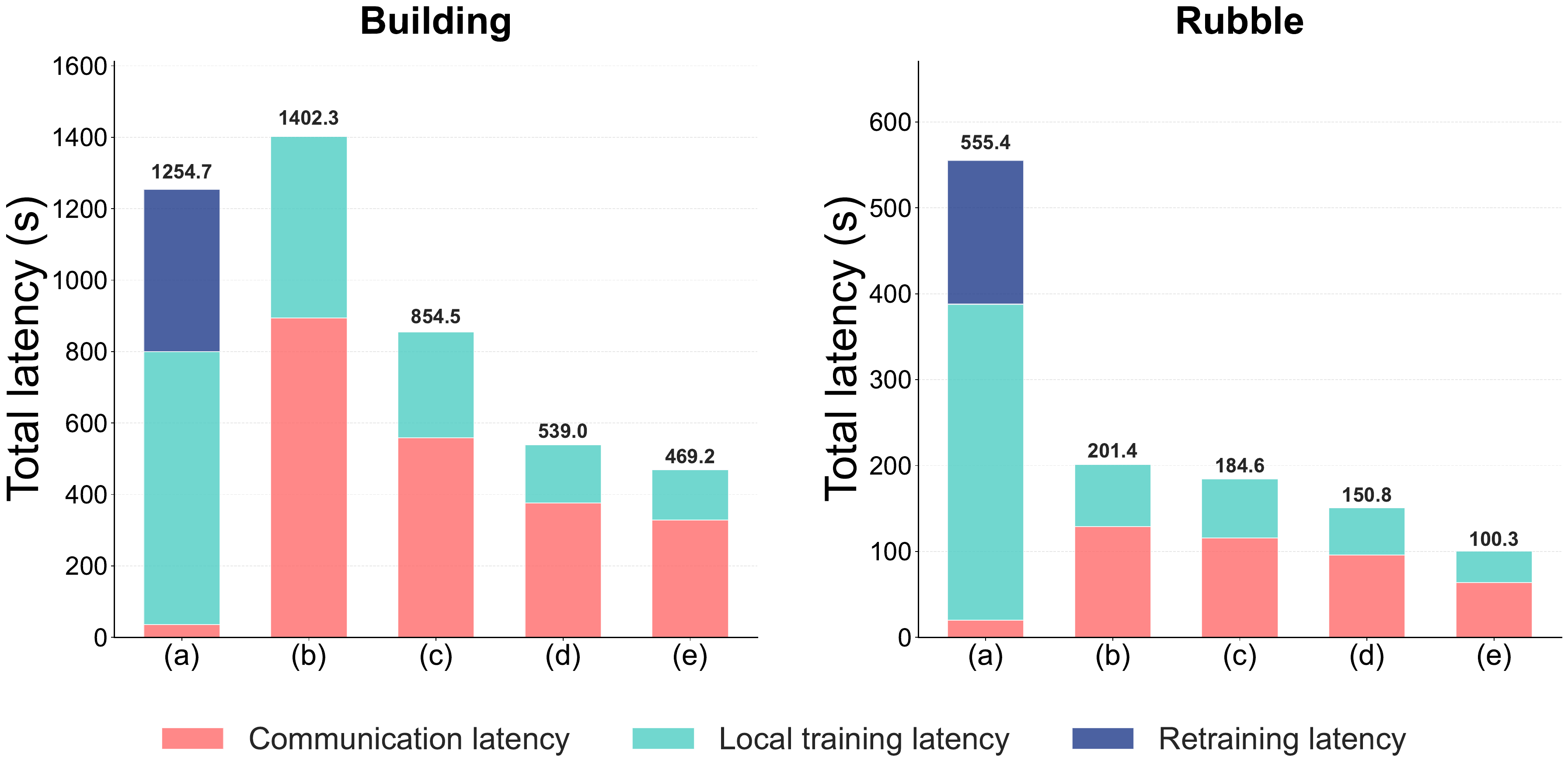} 
  \caption{Comparison of total latency required to first reach a PSNR of 17 dB in the \emph{Building} scene and 19 dB in the \emph{Rubble} scene. Each bar is decomposed into communication latency, local training latency, and retraining latency (if applicable). Schemes (a)–(d) correspond to benchmark methods: (a) Fed3D-GS, (b) full model training design, (c) full memory load with recovery design, and (d)  lightweighting only design, while (e) denotes the proposed design.}
  \label{comparison_chat}
  \vspace{-0.2cm}
\end{figure}

Fig.~\ref{comparison_chat} presents the total latency required to achieve a PSNR of 17 dB on the \emph{Building} scene and 19 dB on the \emph{Rubble} scene. It is observed that the proposed design consistently achieves the lowest latency in both scenes, demonstrating the highest training efficiency. Among the benchmark schemes, Fed3D-GS attains the smallest communication latency owing to its one-shot upload of local models, but suffers from a high retraining latency because it relies on server computation resource to refine the global model. The full model training design incurs the highest communication and second-highest local training latency, since each device repeatedly trains and uploads the complete model in every communication round. The full memory load with recovery design reduces the number of Gaussian points compared to the full model design, thereby slightly lowering both local training and communication latency. The lightweighting only design adopts an optimal strategy to select the number of Gaussian points, resulting in local training latency comparable to our design. However, since it directly merges local models without using recovery mechanism, the global model size increases rapidly over rounds, which leads to higher downlink communication overhead. In comparison, the proposed design effectively lowers both local training and communication latency, and thus achieves the fastest convergence.
\begin{figure}[htbp!]
  \centering
  \includegraphics[width=0.55\linewidth]{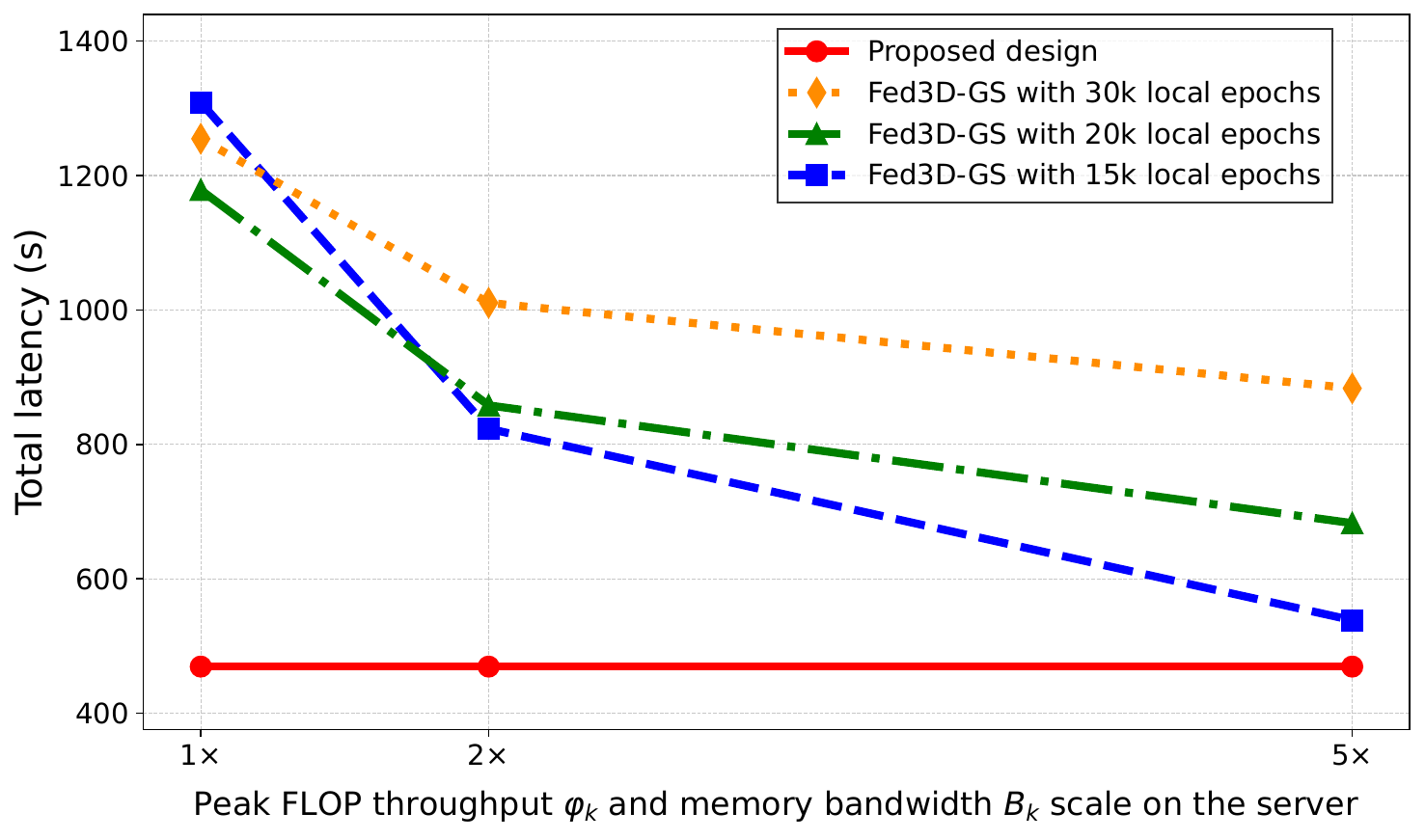} 
  \caption{Comparison of the proposed method and Fed3D-GS (with different numbers of local training epochs) in terms of total latency w.r.t. varying server computation resource.}
  \label{comparison_latency_fed3dgs}
  \vspace{-0.8em}
\end{figure}

Fig.~\ref{comparison_latency_fed3dgs} shows the total latency required to reach a target PSNR of $17~\text{dB}$ in the \emph{building} scene under different peak FLOP throughput $\varphi_k$ and memory bandwidth $B_k$ in \eqref{splat_latency} and \eqref{render_latency}. The scale factors $1 \times, 2 \times$, and $5 \times$ indicate that both $\varphi_k$ and $B_k$ of the server are jointly multiplied by the same factor. It is observed that the proposed design achieves consistently lower latency. In particular, the latency achieved by the proposed design is invariant when the server capability increases. This is because our proposed design fully exploits distributed computation at edge devices for training instead of relying on the server. In contrast, the total latency achieved by Fed3D-GS decreases when the server becomes stronger, since the retraining latency shrinks with increased computational resources. In addition, a stronger server allows Fed3D-GS to reduce the total latency by decreasing the local epochs and relying more on centralized retraining. For instance, when the number of local epochs is reduced from 30k to 20k, the local latency decreases, and although the uploaded models are degraded and require longer retraining, this additional overhead can be compensated by enhanced server resources, leading to a lower total latency. However, when the local epochs are reduced to 15k and below, the uploaded local model becomes too poor to support effective retraining, causing the retraining time to grow rapidly and the total latency to increase again. In particular, when local epochs are very small, the target PSNR cannot be achieved even with the strongest server setting, such that the training becomes infeasible. This demonstrates that our proposed design maintains its effectiveness and advantage in reducing latency when server computation resource varies.

\subsubsection{Visualization Results}

\begin{figure*}[htbp!]
    \centering
    \small
    \begin{minipage}[t]{0.235\textwidth}
        \centering
        \includegraphics[width=0.72\linewidth]{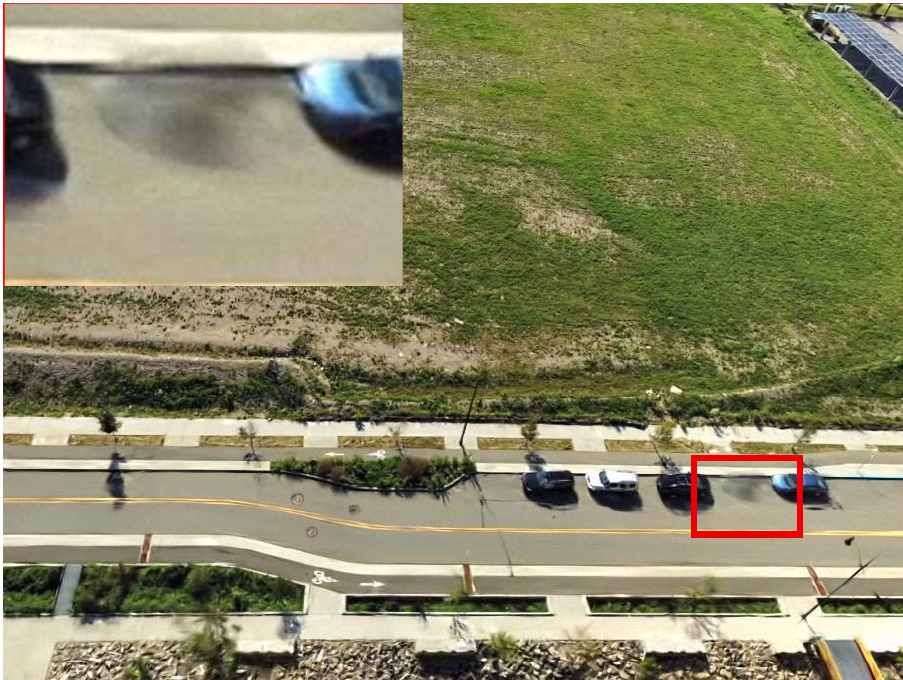} \\[0mm]
        \includegraphics[width=0.72\linewidth]{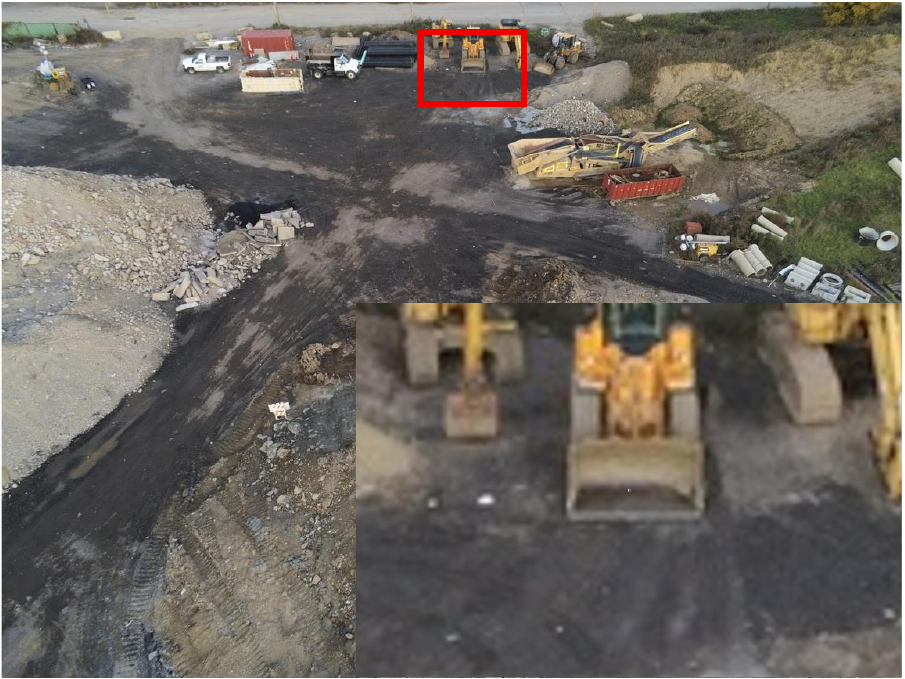} \\[0mm]
        \includegraphics[width=0.72\linewidth]{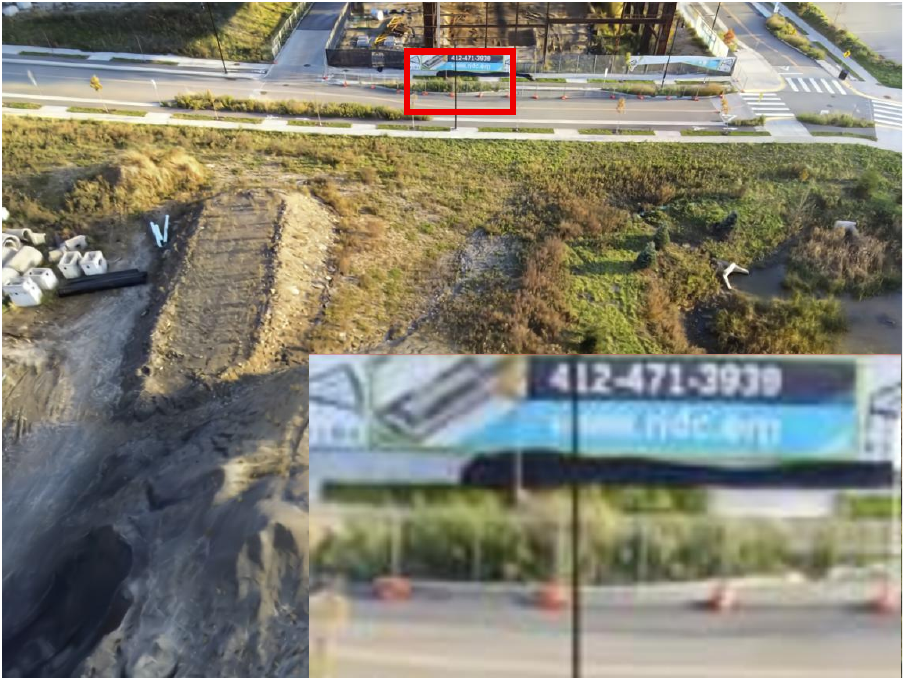} \\[0mm]
        {{3D-GS}}
    \end{minipage}
    \hfill
    \begin{minipage}[t]{0.235\textwidth}
        \centering
        \includegraphics[width=0.72\linewidth]{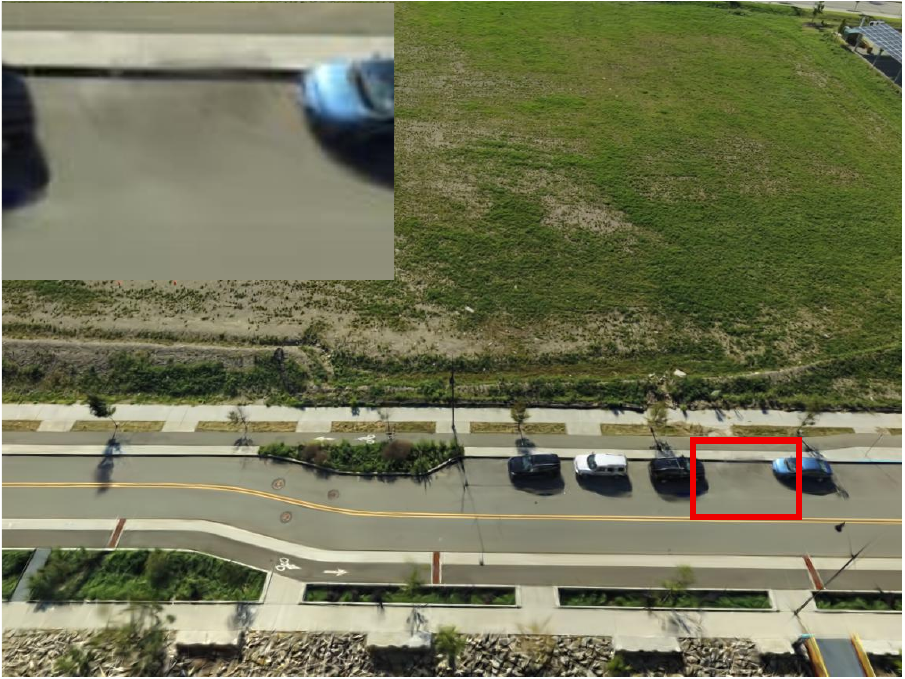} \\[0mm]
        \includegraphics[width=0.72\linewidth]{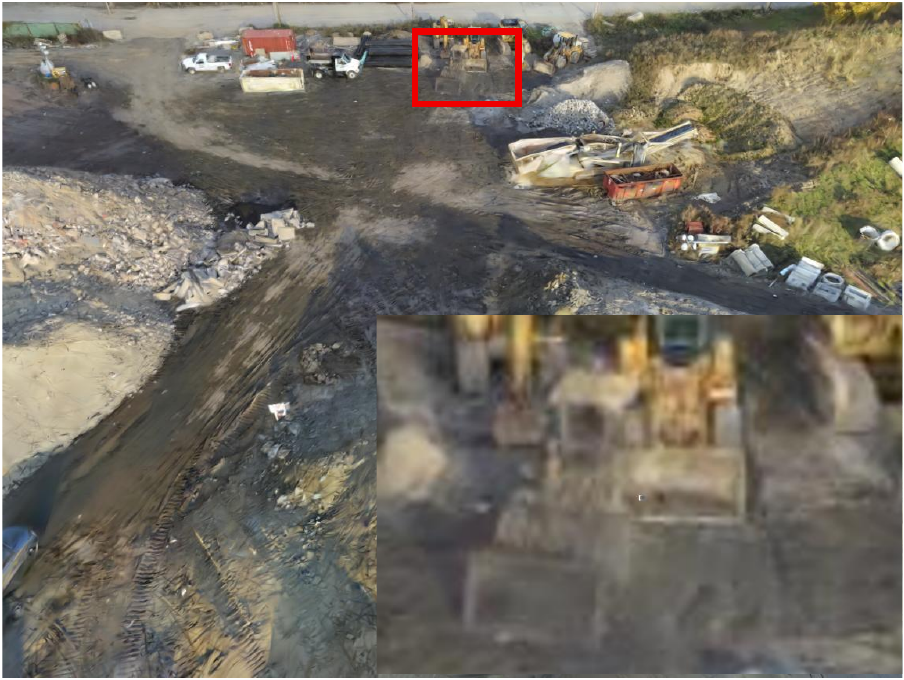} \\[0mm]
        \includegraphics[width=0.72\linewidth]{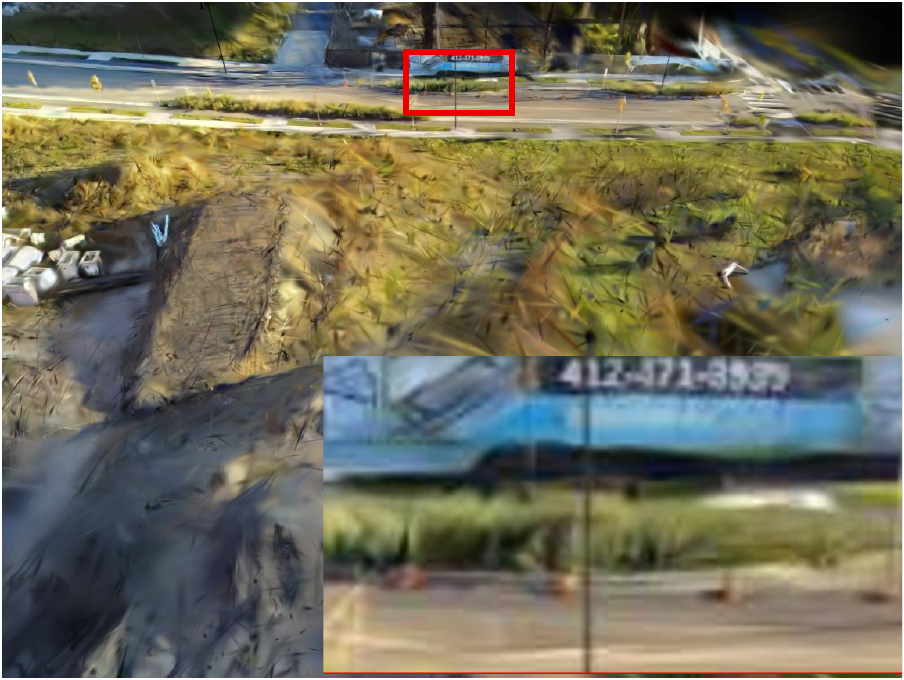} \\[0mm]
        {{Fed3D-GS}}
    \end{minipage}
    \hfill
    \begin{minipage}[t]{0.235\textwidth}
        \centering
        \includegraphics[width=0.72\linewidth]{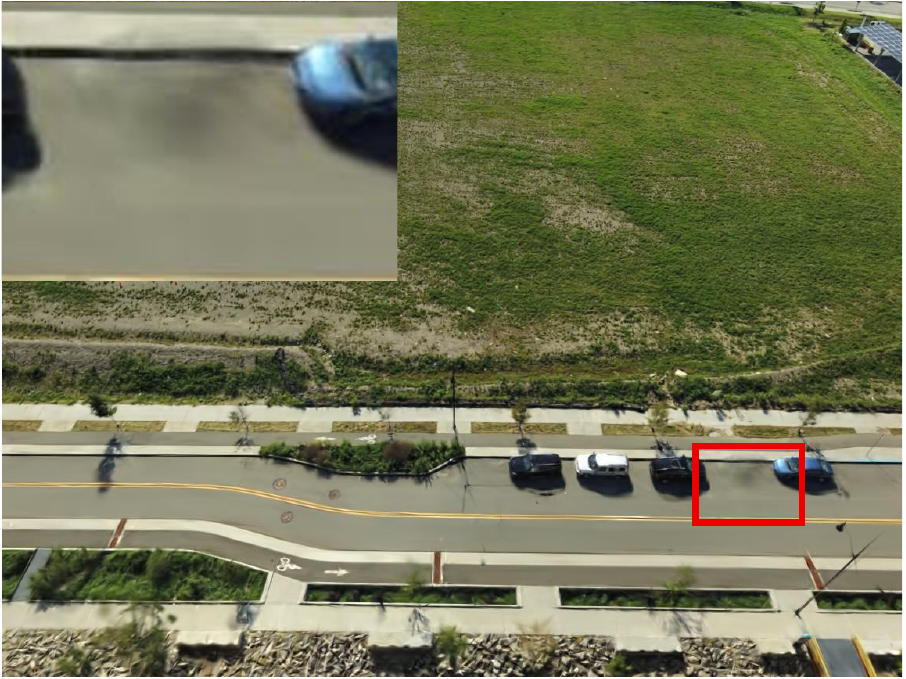} \\[0mm]
        \includegraphics[width=0.72\linewidth]{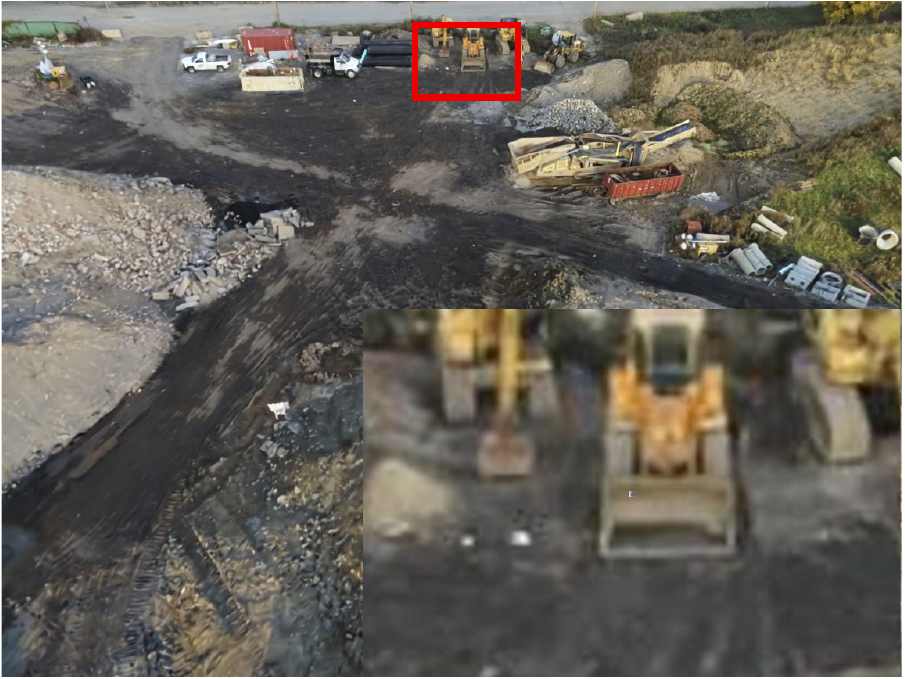} \\[0mm]
        \includegraphics[width=0.72\linewidth]{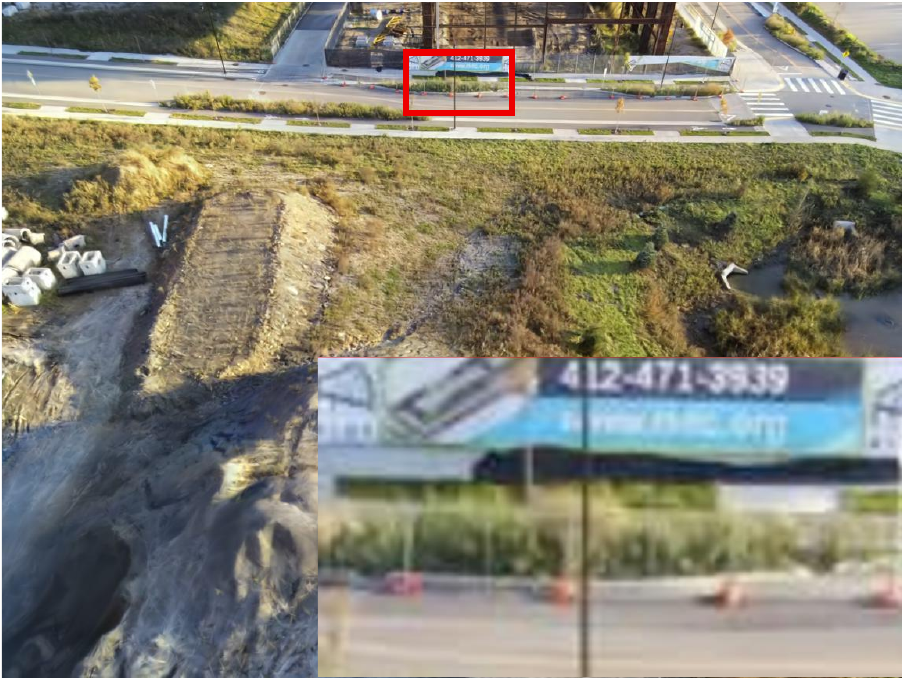} \\[0mm]
        {{Lightweighting only}}
    \end{minipage}
    \hfill
    \begin{minipage}[t]{0.235\textwidth}
        \centering
        \includegraphics[width=0.72\linewidth]{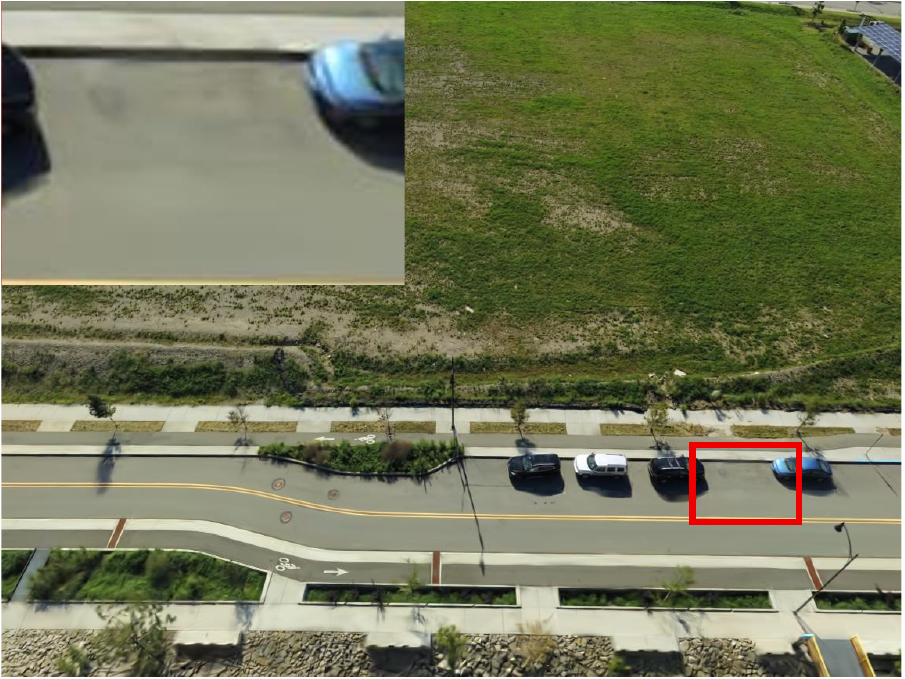} \\[0mm]
        \includegraphics[width=0.72\linewidth]{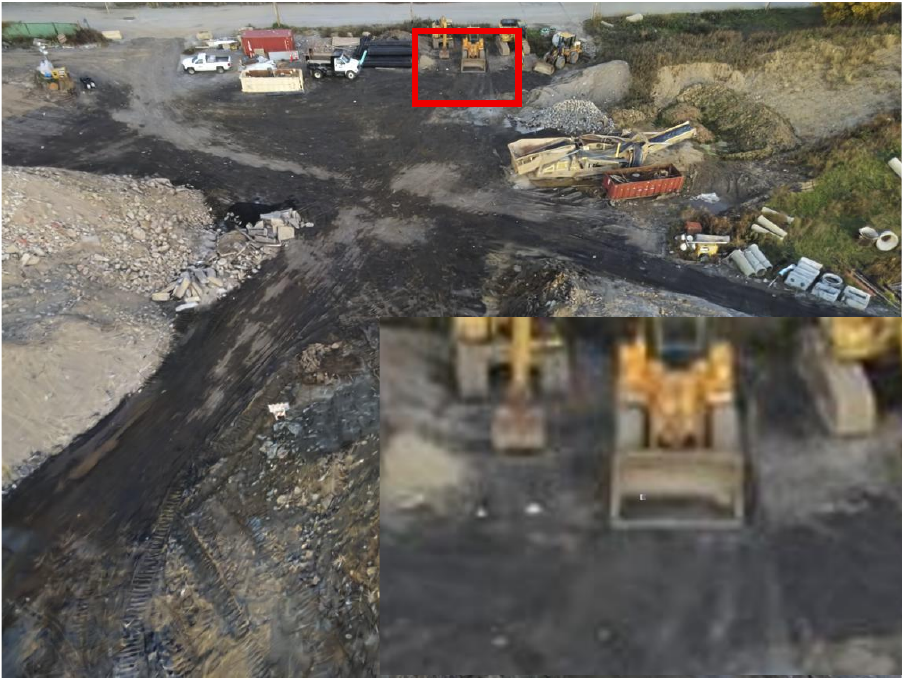} \\[0mm]
        \includegraphics[width=0.72\linewidth]{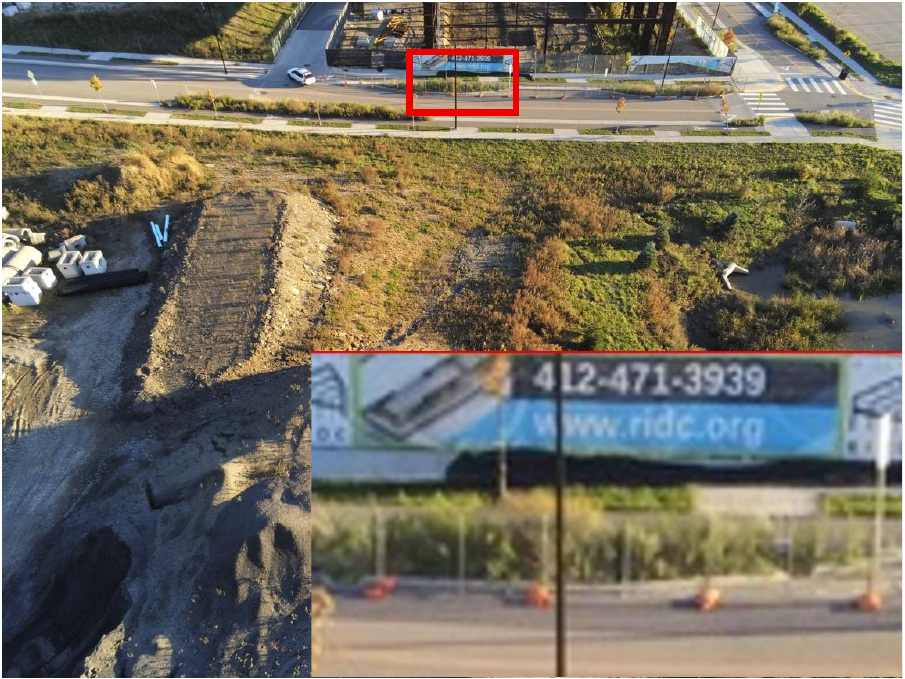} \\[0mm]
        {{Our proposed design}}
    \end{minipage}

    \caption{Qualitative comparison of novel view rendering among 3D-GS, Fed3D-GS, the lightweighting only design, and the proposed method. Our proposed design efficiently suppresses visual artifacts and achieves better reconstruction quality compared with the other benchmark schemes. The first row shows the results on the \textit{building} scene, while the second and third rows show the results on the \textit{rubble} scene.}
    \label{qualitative_mill19}
\vspace{-1.8em}
\end{figure*}

{Fig.~\ref{qualitative_mill19} compares novel view rendering results of our method against centralized 3D-GS, Fed3D-GS, and the lightweighting only design.} The evaluation is conducted using camera parameters corresponding to unseen viewpoints that are excluded from the training sets. {It is observed that the proposed method achieves visual fidelity comparable to centralized 3D-GS, although the latter is trained centrally using all views. As shown in the third row, our method reconstructs the scene structure more completely and preserves finer details than Fed3D-GS and the lightweighting only design, while remaining competitive with centralized 3D-GS.} In addition, our method provides more accurate rendering in fine-grained regions, e.g., it better preserves the shadow contours along the sidewalk in the \emph{Building} scene. More importantly, this result also demonstrates the impact of our proposed structure-consistent recovery mechanism. Both the Fed3D-GS and the lightweighting only design suffer from visual artifacts such as floaters, which are highlighted with red boxes in the figure. In comparison, our method effectively suppresses these artifacts. This improvement is attributed to the proposed recovery mechanism, which ensures that each Gaussian point retains the opportunity to participate in parameter averaging during global model aggregation, rather than being permanently excluded after pruning in model lightweighting. By integrating contributions from diverse device viewpoints, the recovered 3D-GS model avoids overfitting to the local dataset and thus mitigates the emergence of floaters when rendering from novel views.

\vspace{-0.08cm}
\section{Conclusion}

This paper proposed a resource-efficient federated 3D-GS framework for large-scale scene reconstruction at the wireless edge under constrained memory, computation, and communication resources. We proposed a latency- and memory-aware lightweighting mechanism to adaptively select and prune Gaussian points based on a novel importance-to-latency ratio, maximizing training efficiency adhering to resource limitations at each device. Additionally, a structure-consistent recovery mechanism was introduced to enable effective model aggregation. Experiments on outdoor large-scale scenes demonstrated that our framework significantly accelerated convergence, reduced overall latency, and preserved rendering fidelity compared to benchmark schemes.

\appendix[Proof of Theorem \ref{theorem}] \label{appendix_1}
For notational simplicity, we omit the communication round index $n$ and device index $k$ in the following proof. We focus on a subset of candidate Gaussian point counts defined in \eqref{subset}. By letting $\quad S(U)=\sum_{j=1}^U \tilde{w}_j, \quad U \in \mathcal{B}_b$, the optimization objective is defined as  $ \rho(U)=\frac{S(U)}{T(U)}$. The latency function $T(U)$ can be formulated by $T(U)=\alpha_0+\alpha_1 U+\alpha_2\left\lceil\frac{U}{\beta}\right\rceil$, where $\alpha_0=M\left\lceil\frac{|\boldsymbol{I}|}{\beta_k}\right\rceil \phi_1 \frac{\tilde{N}_{\text{FLOPs}}}{\tilde{\epsilon} \tilde{N}^{\text{roofline}}}+\frac{z_{\text{para}} q_{\text{para}} \bar{U}}{r^{\text{d}}}>0$, $\alpha_1 = M\left\lceil\frac{|\boldsymbol{I}|}{\beta_k}\right\rceil\left(\frac{\phi_2}{|\boldsymbol{I}|}\right) \frac{\tilde{N}_{\text{FLOPs}}}{\tilde{\epsilon} \tilde{N}^{\text{roofline}}}+\frac{z_{\text{para}} q_{\text{para}}+\operatorname{dim}(\boldsymbol{x}) q_{\text{opt}}}{r^{\text{u}}}>0$, and $\alpha_2 = M \frac{N_{\text{FLOPs}}}{\epsilon N^{\text{roofline}}}>0$. Then, we analyze how the objective $\rho(U)$ changes when $U$ increases. For any $ U\in \mathcal{B}_b$ with $U+1\in\mathcal{B}_b$, we have $\Delta T(U)=T(U+1)-T(U) = \alpha_1+\alpha_2\chi(U),$
where $\chi(U)=\left\lceil\frac{U+1}{\beta}\right\rceil-\left\lceil\frac{U}{\beta}\right\rceil$. As such, $\Delta \rho(U)$ is given by
\begin{align}
    \Delta \rho(U)=\rho(U+1)-\rho(U) =\frac{\tilde{w}_{U+1} T(U)-S(U) \Delta T(U)}{T(U)(T(U)+\Delta T(U))}.
\end{align}
It is observed that the denominator term $T(U)(T(U)+\Delta T(U)) >0$ due to the fact that $\Delta T(U)$ and $T(U) >0$. Consequently, the sign of $ \Delta \rho(U)$ is fully determined by the numerator, which is denoted as $\Lambda(U)=\tilde{w}_{U+1} T(U)-S(U) \Delta T(U)$.

{Next, we analyze the monotonicity of $\Lambda(U)$. For any
$U\in\mathbb{Z}_{+}$ satisfying $b\beta+1
\leq U
\leq
\min\{(b+1)\beta-3,\widehat{U}-2\},$
we have $U,U+1,U+2\in\mathcal{B}_b$. We have
\begin{align}
\Lambda(U+1)-\Lambda(U)
&=
(\widetilde{w}_{U+2}-\widetilde{w}_{U+1})T(U+1)
\nonumber\\
&\quad
-S(U+1)
\big(\Delta T(U+1)-\Delta T(U)\big).
\end{align}
For any $U$ in the above range,
$\left\lceil\frac{U}{\beta}\right\rceil
=
\left\lceil\frac{U+1}{\beta}\right\rceil
=
\left\lceil\frac{U+2}{\beta}\right\rceil
=b+1.$
Therefore, $\Delta T(U+1)-\Delta T(U)=0.$
Together with
$\widetilde{w}_{U+2}\leq\widetilde{w}_{U+1}$ and
$T(U+1)>0$, we have $\Lambda(U+1)-\Lambda(U)\leq0.$
Hence, $\Lambda(U)$ is non-increasing over its valid indices
within each block $\mathcal{B}_b$. Accordingly,
$\Delta\rho(U)$ can change its sign at most once from
non-negative to non-positive. This completes the proof.}
\vspace{-0.15cm}

\bibliographystyle{IEEEtran}

\bibliography{IEEEabrv,reference}
\vspace{12pt}

\end{document}